\documentclass[conference]{IEEEtran}
\IEEEoverridecommandlockouts
\usepackage{cite}
\usepackage{amsmath,amssymb,amsfonts}
\usepackage{algorithmic}
\usepackage{graphicx}
\usepackage{textcomp}
\usepackage{array}
\usepackage{color}
\usepackage{multirow}
\usepackage{booktabs}
\usepackage{makecell}
\usepackage{supertabular}
\usepackage{afterpage}
\usepackage[numbers,sort&compress]{natbib}
\usepackage{url}

\usepackage[colorlinks,
            linkcolor=black,
            anchorcolor=black,
            citecolor=black,
            urlcolor=black
            ]{hyperref}

\begin{document}

\title{A Survey on Face Data Augmentation\\
}

\author{\IEEEauthorblockN{Xiang Wang}
\IEEEauthorblockA{\textit{CloudMinds Technologies Inc.} \\
xiang.wang@cloudminds.com}
\and
\IEEEauthorblockN{Kai Wang}
\IEEEauthorblockA{\textit{CloudMinds Technologies Inc.} \\
kai.wang@cloudminds.com}
\and
\IEEEauthorblockN{Shiguo Lian}
\IEEEauthorblockA{\textit{CloudMinds Technologies Inc.} \\
scott.lian@cloudminds.com}
}

\maketitle

\begin{abstract}
The quality and size of training set have great impact on the results of deep learning-based face related tasks. However, collecting and labeling adequate samples with high quality and balanced distributions still remains a laborious and expensive work, and various data augmentation techniques have thus been widely used to enrich the training dataset. In this paper, we systematically review the existing works of face data augmentation from the perspectives of the transformation types and methods, with the state-of-the-art approaches involved. Among all these approaches, we put the emphasis on the deep learning-based works, especially the generative adversarial networks which have been recognized as more powerful and effective tools in recent years. We present their principles, discuss the results and show their applications as well as limitations. Different evaluation metrics for evaluating these approaches are also introduced. We point out the challenges and opportunities in the field of face data augmentation, and provide brief yet insightful discussions.

\end{abstract}


\section{Introduction}

Human face plays a key role in personal identification, emotional expression and interaction. In the last decades, a number of popular research subjects related to face have grown up in the community of computer vision, such as facial landmark detection, face alignment, face recognition, face verification, emotion classification, etc. As well as many other computer vision subjects, face study has shifted from engineering features by hand to using deep learning approaches in recent years. In these methods, data plays a central role, as the performance of the deep neural network heavily depends on the amount and quality of the training data.

The remarkable work by Facebook~\cite{taigman2014deepface} and Google~\cite{schroff2015facenet} demonstrated the effectiveness of large-scale datasets on obtaining high-quality trained model, and revealed that deep learning strongly relies on large and complex training sets to generalize well in unconstrained settings. This close relationship of the data and the model effectiveness has been further verified in~\cite{hestness2017deep}. However, collecting and labeling a large quantity of real samples is widely recognized as laborious, expensive and error-prone, and existing datasets are still lack of variations comparing to the samples in the real world.

\begin{figure}[htbp]
\centerline{\includegraphics[width=0.45\textwidth]{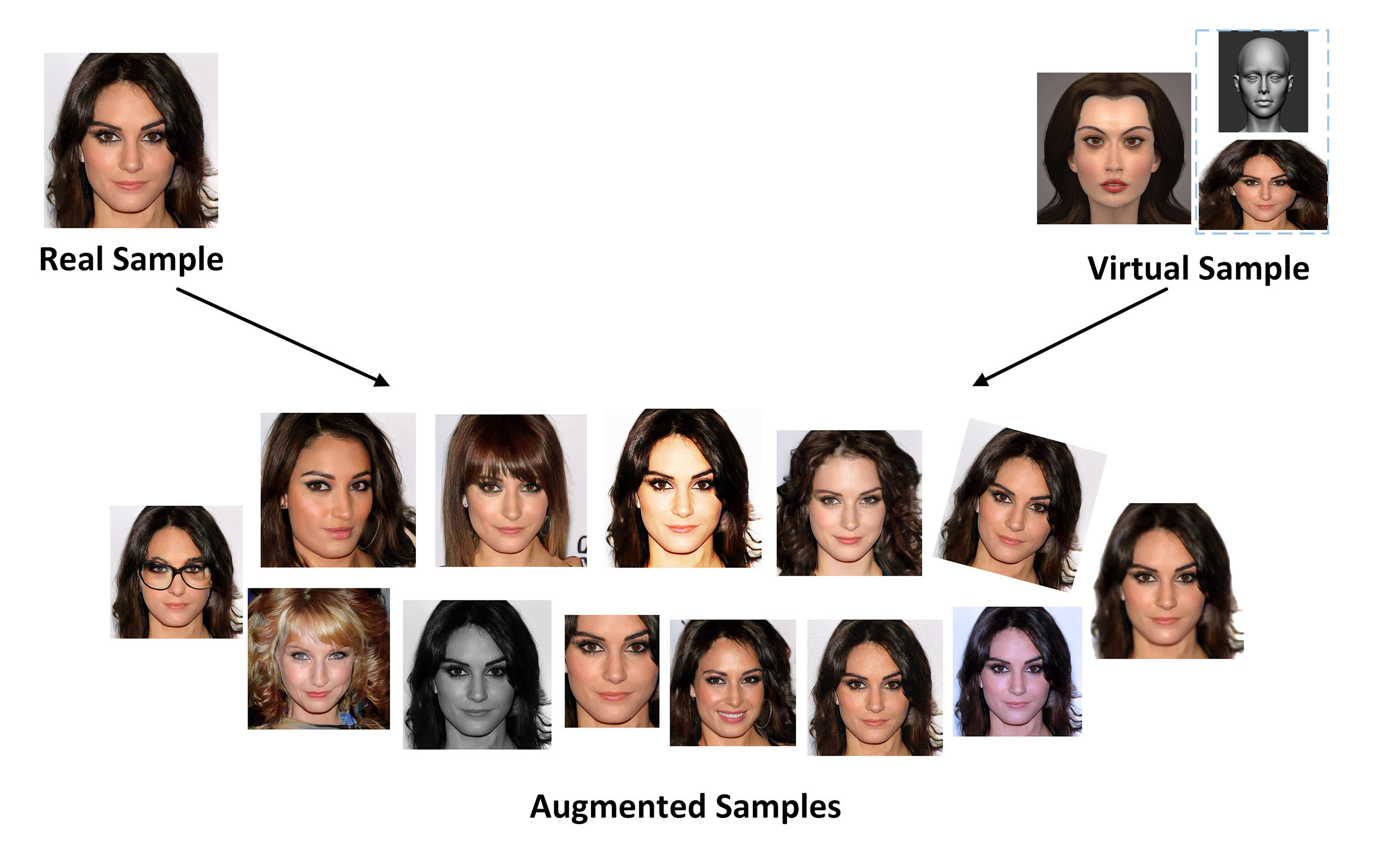}}
\caption{Schematic diagram of face data augmentation. The real and augmented samples were generated by TL-GAN (transparent latent-space Generative Adversarial Network) \cite{guan2018tlgan}.}
\label{AugFig}
\end{figure}

To compensate the insufficient facial training data, data augmentation provides an effective alternative, which we call "face data augmentation". It is a technology to enlarge the data size of training or testing by transforming collected real face samples or simulated virtual face samples. Fig.~\ref{AugFig} shows a schematic diagram of face data augmentation, which is our focus in this paper.

Assuming the original dataset is $\mathcal{S}$, face data augmentation can be represented by the following mapping:
\begin{equation}\label{eq:map}
\phi: \mathcal{S} {\mapsto} \mathcal{T},
\end{equation}

where $\mathcal{T}$ is the augmented set of $\mathcal{S}$. Then the dataset is enlarged as the union of the original set and the augmented set:
\begin{equation}
\mathcal{S'} = \mathcal{S} \cup \mathcal{T}.
\end{equation}

The direct motivation for face data augmentation is to overcome the limitation of existing data. Insufficient data amount or unbalanced data distribution will cause overfitting and over-parameterization problems, leading to an obvious decrease in the effectiveness of learning result.

Face data augmentation is fundamentally important for improving the performance of neural networks in the following aspects. (1) It is inexpensive to generate a huge number of synthetic data with annotations in comparison to collecting and labeling real data. (2) Synthetic data can be accurate, so it has groundtruth by nature. (3) If controllable generation method is adopted, faces with specific features and attributes can be obtained. (4) Face data augmentation has some special advantages, such as generating faces without self-occlusion \cite{feng2015cascaded} and balanced dataset with more intra-class variations \cite{masi2016we}.

At the same time, face data augmentation has some limitations. (1) The generated data lack realistic variations in appearance, such as variations in lighting, make-up, skin color, occlusion and sophisticated background, which means the synthetic data domain has different distribution to real data domain. That is why some researchers use domain adaption and transfer learning techniques to improve the utility of synthetic data \cite{liu2018feature, hong2017sspp}. (2) The creation of high-quality synthetic data is challenging. Most generated face images lack facial details, and the resolution is not high. Furthermore, some other problems are still under study, such as identity preserving and large-pose variation.

This paper aims to give an epitome of face data augmentation, especially on what can face data augmentation do and how to augment the existing face data, including both the traditional methods and the up-to-date approaches. In addition, we thoroughly discuss the challenges and open problems of face data augmentation for further research. Data augmentation has overlap with data synthesis/generation, but differs with them in the point that the augmented data is generated based on existing data. In fact, many data synthesis techniques can be applied to data augmentation. Although some works were not designed for data augmentation, we also include them in this survey.

The remainder of the paper is organised as follows. The review of related works is given in Sect. \ref{sec:relwork}. Then the transformation types of face data augmentation are elaborated in Sect. \ref{sec:transtype}, and the commonly used methods are introduced in Sect. \ref{sec:methods}. Sect. \ref{sec:evaluation} provides a description of the evaluation metrics. Sect. \ref{sec:challenges} presents some challenges and potential research directions. Discussions and conclusion are given in the last two sections respectively.

\section{Related Work}\label{sec:relwork}

This section reviews the existing works that have in-depth analysis and evaluation on data augmentation techniques. Masi et al.~\cite{masi2016we} discussed the necessity of collecting huge numbers of face images for effective face recognition, and proposed a synthesizing method to enrich the existing dataset by introducing face appearance variations for pose, shape and expression. Lv et al.~\cite{lv2017data} presented five data augmentation methods for face images, including landmark perturbation, hairstyle synthesis, glasses synthesis, poses synthesis and illumination synthesis. They tested these methods on different datasets for face recognition, and compared their performance. Taylor et al.~\cite{taylor2017improving} demonstrated the effectiveness of using basic geometric and photometric data augmentation schemes like cropping, rotation, etc., to help researchers find the most appropriate choice for their dataset. Wang et al. \cite{wang2017effectiveness} compared traditional transformation methods with GANs (Generative Adversarial Networks) to the problem of data augmentation in image classification. In addition, they proposed a network-based augmentation method to learn augmentations that best improve the classifier in the context of generic images, not face images. Kortylewski et al.~\cite{kortylewski2018training} explored the ability of data augmentation to train deep face recognition systems with an off-the-shelf face recognition software and fully synthetic face images varying in pose, illumination and background. Their expriment demonstrated that synthetic data with strong variations performed well across different datasets even without dataset adaptation, and the domain gap between the real and the synthetic could be closed when using synthetic data for pre-training followed by fine-tuning. Li et al.~\cite{li2018survey} reviewed the research progress in the field of virtual sample generation for face recognition. They categorized the existing methods into three groups: construction of virtual face images based on face structure, perturbation and distribution function, and sample viewpoint. Compared to the existing works, our survey covers a wider range of face data augmentation methods, and contains the up-to-date researches. We introduce these researches in intuitive presentation level and deep method level.

\section{Transformation Types}\label{sec:transtype}

In this section, we elaborate the transformation types, including the generic and face specific transformations, for producing the augmented samples \(\mathcal{T}\) in Eq.~\ref{eq:map}. The applications of some methods go beyond face data augmentation to other learning-based computer vision tasks. Usually, the generic methods transform the entire image and ignore high-level contents, while face specific methods focus on face components or attributes and are capable of transforming age, makeup, hairstyle, etc. Table I shows an overview of the commonly used face data transformations.

\begin{table}[htbp]
\caption{An overview of transformation types}
\begin{center}
\begin{tabular}{|p{1.8cm}|cc|}
\hline
\multirow{2}*{\begin{minipage}[c][1.4cm][c]{0.1\textwidth}
\hspace{1mm} \centering Generic\newline Transformation
\end{minipage}}
&Geometric&\begin{minipage}{0.23\textwidth}
\vspace{1mm} \includegraphics[width=0.95\textwidth]{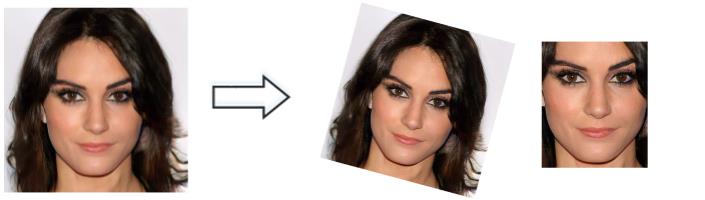} \vspace{1mm}
\end{minipage} \\
&Photometric&\begin{minipage}{0.23\textwidth}
\vspace{1mm} \includegraphics[width=0.95\textwidth]{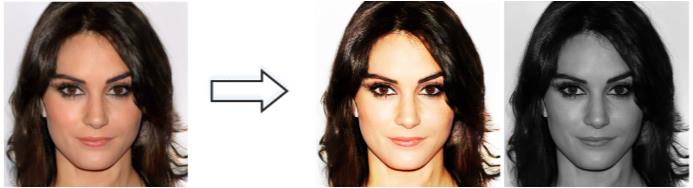} \vspace{1mm}
\end{minipage} \\
\hline
\multirow{3}*{\begin{minipage}[c][3.2cm][c]{0.1\textwidth}
\hspace{1mm} \centering Component\newline Transformation
\end{minipage}}
&Hairstyle&\begin{minipage}{0.23\textwidth}
\vspace{1mm} \includegraphics[width=0.95\textwidth]{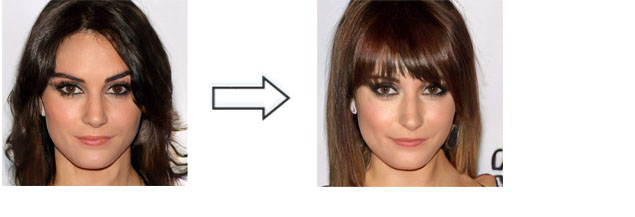} \vspace{1mm}
\end{minipage} \\
&Makeup&\begin{minipage}{0.23\textwidth}
\vspace{1mm} \includegraphics[width=0.90\textwidth]{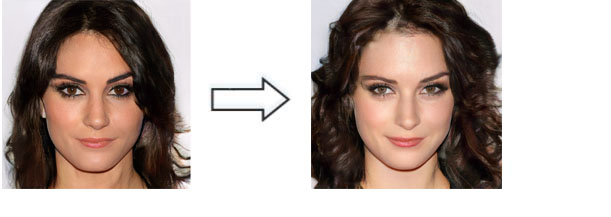} \vspace{1mm}
\end{minipage} \\
&Accessory&\begin{minipage}{0.23\textwidth}
\vspace{1mm} \includegraphics[width=0.95\textwidth]{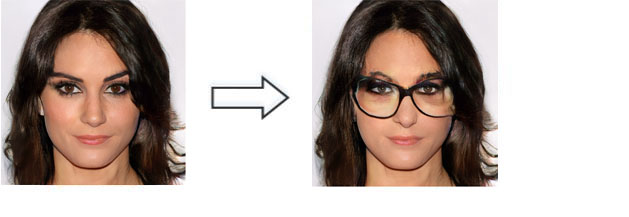} \vspace{1mm}
\end{minipage} \\
\hline
\multirow{3}*{\begin{minipage}[c][3.2cm][c]{0.1\textwidth}
\hspace{1mm} \centering Attribute\newline Transformation
\end{minipage}}
&Pose&\begin{minipage}{0.23\textwidth}
\vspace{1mm} \includegraphics[width=0.95\textwidth]{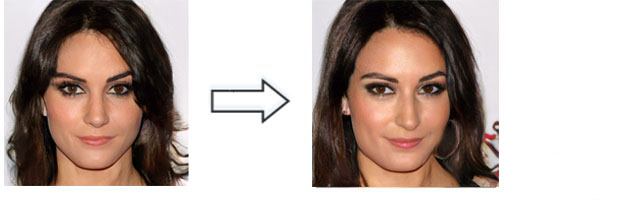} \vspace{1mm}
\end{minipage} \\
&Expression&\begin{minipage}{0.23\textwidth}
\vspace{1mm} \includegraphics[width=0.95\textwidth]{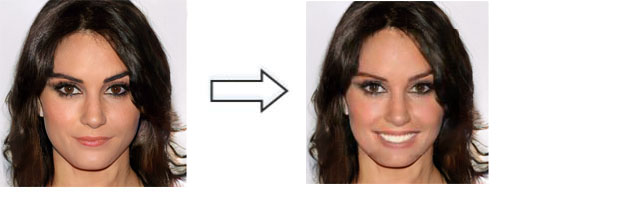} \vspace{1mm}
\end{minipage} \\
&Age&\begin{minipage}{0.23\textwidth}
\vspace{1mm} \includegraphics[width=0.95\textwidth]{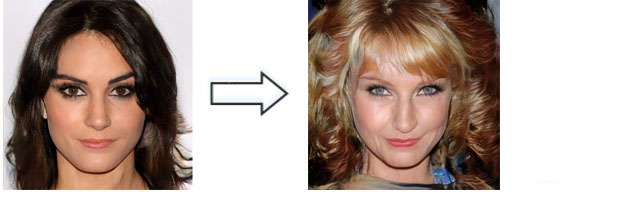} \vspace{1mm}
\end{minipage} \\
\hline
\end{tabular}
\label{tab1}
\end{center}
\end{table}

\subsection{Geometric and Photometric Transformation}\label{sec:transtype:gpt}

The generic data augmentation techniques can be divided into two categories: geometric transformation and photometric transformation. These methods have been adapted to various learning-based computer vision tasks.

\begin{figure}[htbp]
\centerline{\includegraphics[width=0.47\textwidth]{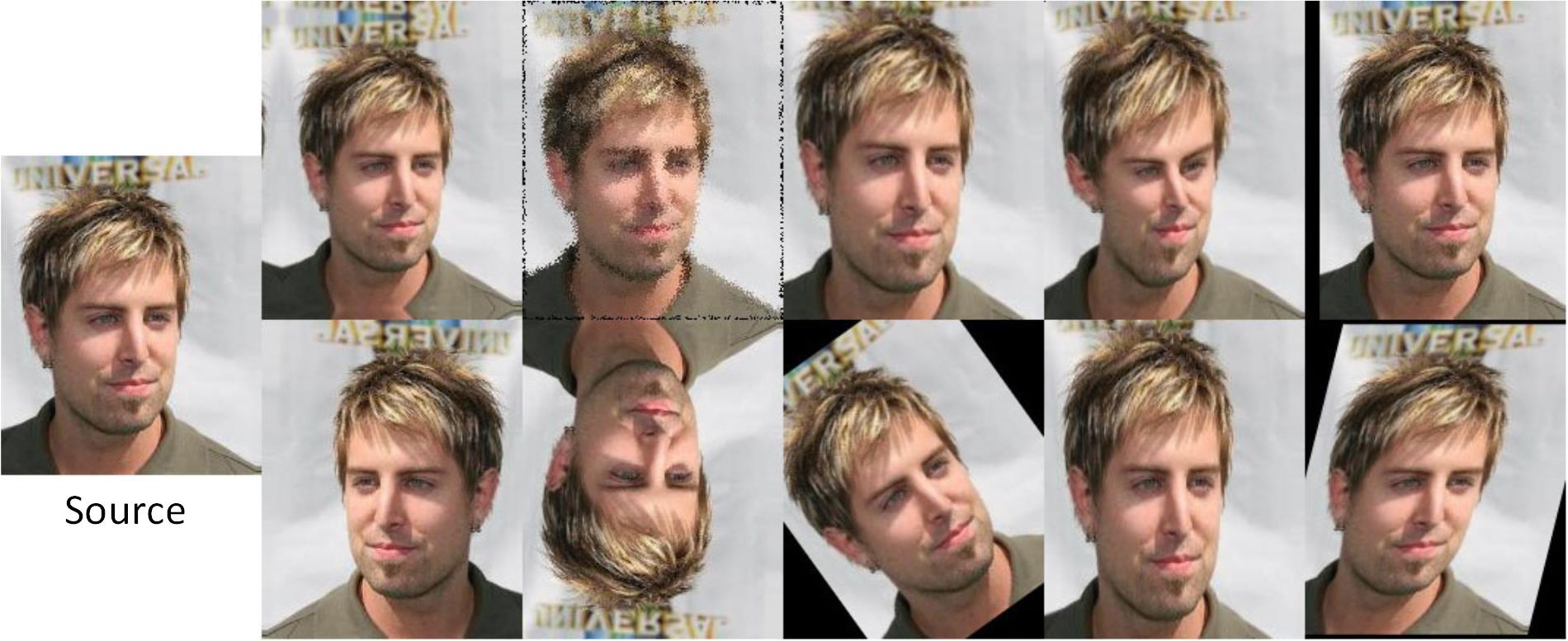}}
\caption{Geometric transformation examples created by imgaug \cite{jung2017imgaug}. The source image is from CelebA dataset\cite{liu2015deep}. From left to right and from top to bottom, the transformations are crop\&pad, elastic distortion, scale, piecewise affine, translate, horizontal flip, vertical flip, rotate, perspective transformation, and shear.}
\label{GeoTrans}
\end{figure}

Geometric transformation alters the geometry of an image by transferring image pixel values to new positions. This kind of transformation includes translation, rotation, reflection, flipping, zoomming, scaling, cropping, padding, perspective transformation, elastic distortion, lens distortion, mirroring, etc. Some examples are illustrated in Fig. \ref{GeoTrans}, which are created using imgaug--a python library for image augmentation~\cite{jung2017imgaug}.

\begin{figure}[htbp]
\centerline{\includegraphics[width=0.47\textwidth]{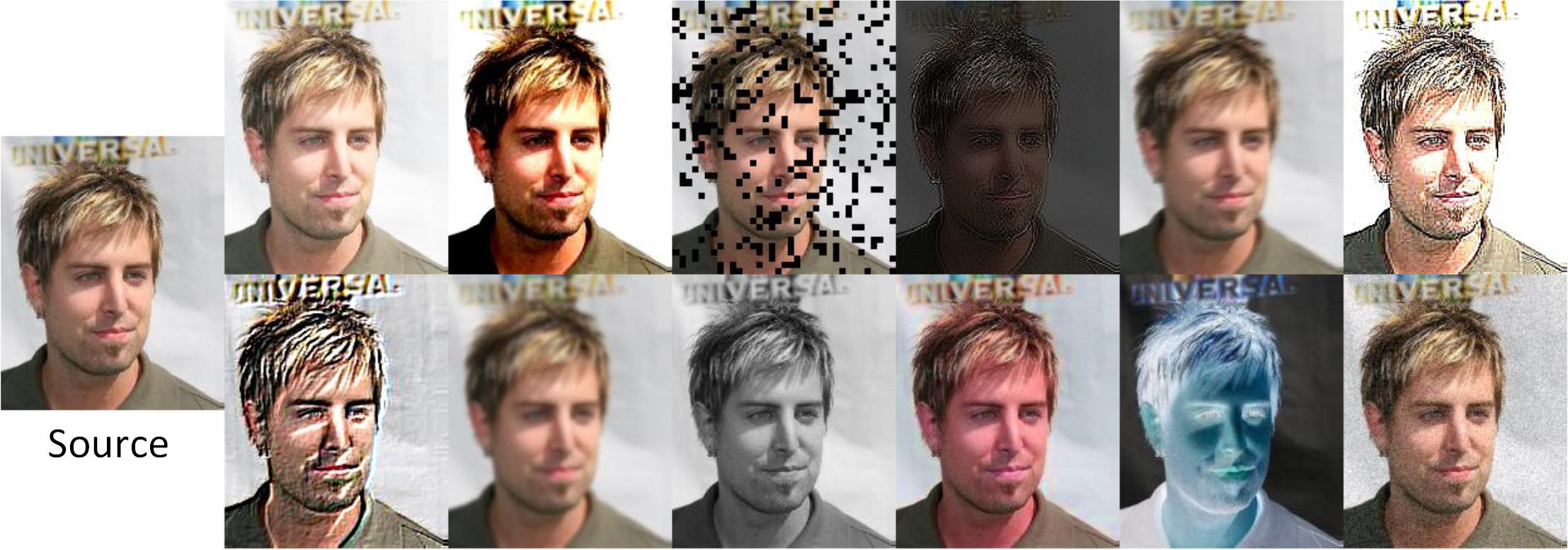}}
\caption{Photometric transformation examples created by imgaug \cite{jung2017imgaug}. The source image is from CelebA dataset\cite{liu2015deep}. From left to right and from top to bottom, the transformations are brightness change, contrast change, coarse dropout, edge detection, motion blur, sharpen, emboss, Gaussian blur, hue and saturation change, invert, and adding noise.}
\label{PhoTrans}
\end{figure}

Photometric transformation alters the RGB channels by shifting pixel colors to new values, and the main approaches include color jittering, grayscaling, filtering, lighting perturbation, noise adding, vignetting, contrast adjustment, random erasing, etc. The color jittering method includes many different manipulations, such as inverting, adding, decreasing and multiply. The filtering method includes edge enhancement, blurring, sharpening, embossing, etc. Some examples of the photometric transformations are shown in Fig. \ref{PhoTrans}.

Wu et al.~\cite{hsu2018investigating} adopted a series of geometric and photometric transformations to enrich the training dataset and prevent overfitting. \cite{lv2017data} evaluated six geometric and photometric data augmentation methods with baseline for the task of image classification. The six data augmentation methods included flipping, rotating, cropping, color jittering, edge enhancement and fancy PCA (adding multiples of principle components to the images). Their results indicated that data augmentation helped improving the classification performance of Convolutional Neural Network(CNN) in all cases, and the geometric augmentation schemes outperformed the photometric schemes. Mash et al.~\cite{mash2016improved} also benchmarked a variety of augmentation methods, including cropping, rotating, rescaling, polygon occlusion and their combinations, in the context of CNN-based fine-grain aircraft classification. The experimental results showed that flipping and cropping have more obvious improvement on the classifier performance than random scaling and occlusions, which is consistent with the demonstrations in other generic data augmentation works.

\subsection{Hairstyle Transfer}\label{sec:transtype:hair}

Although hair is not an internal component of human face, it affects face detection and recognition due to the occlusion and appearance variation of face it caused. The data augmentation technique is used to generate face images with different hairs in color, shape and bang.

\begin{figure}[htbp]
\centerline{\includegraphics[scale=0.3]{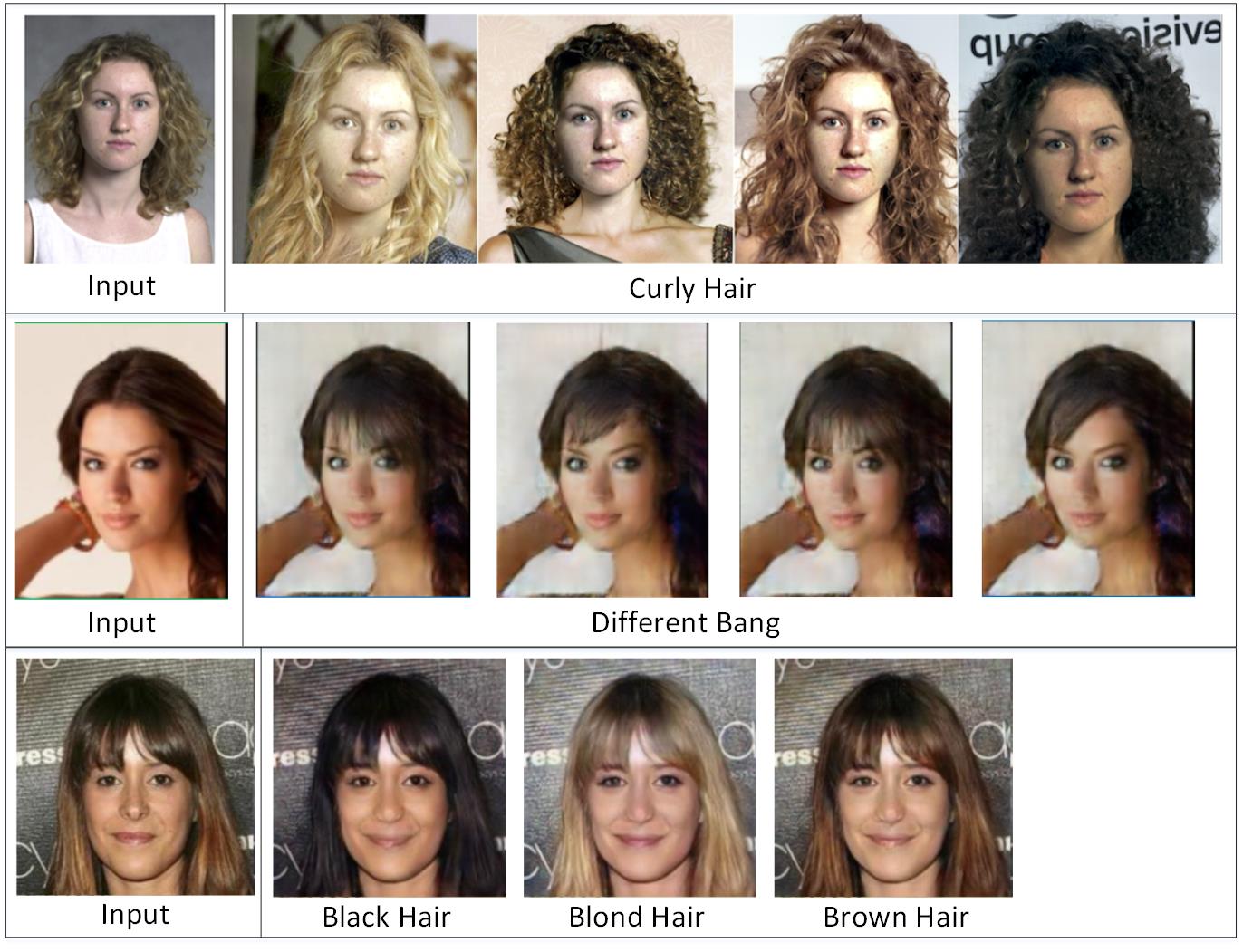}}
\caption{Some examples of hairstyle transfer. The results in the first, second and bottom rows are from \cite{kemelmacher2016transfiguring}, \cite{kim2017unsupervised} and \cite{choi2018stargan} respectively.}
\label{HairTrans}
\end{figure}

Kim et al.~\cite{kim2017learning} transformed hair color using DiscoGAN, which was introduced to discover cross-domain relations with unpaired data. The same function was achiveved by StarGAN \cite{choi2018stargan}, whereas it could perform multi-domain translations using a single model. Besides the color, \cite{kim2017unsupervised} proposed an unsupervised visual attribute transfer using reconfigurable generative adversarial network to change the bang. Lv et al.~\cite{lv2017data} synthesized images with various hairstyles based on hairstyle templates. \cite{kemelmacher2016transfiguring} presented a face synthesis system utilizing an Internet-based compositing method. With one or more photos of a person¡¯s face and a text query ¡°curly hair¡± as input, the system could generate a series of new photos with the input person¡¯s identity and queried appearance. Fig. \ref{HairTrans} presents some examples of hairstyle transfer.

\subsection{Facial Makeup Transfer}\label{sec:transtype:makeup}

While makeup is a ubiquitous way to improve one's facial appearance, it increases the difficulty on accurate face recognition. Therefore, numerous samples with different makeup styles should be provided in the training data to make the algorithm be more robust. Facial makeup transfer aims to shift the makeup style from a given reference to another face while preserving the face identity. Common makeups include foundation, eye linear, eye shadow, lipstick, etc., and their transformations include applying, removal and exchange. Most existing studies on automatic makeup transfer can be classified into two categories: traditional image processing approaches like gradient editing and alpha blending \cite{guo2009digital, oo2016digital, lee2016anew}, and deep learning based methods \cite{alashkar2017examples, li2018beautygan, chang2018pairedcyclegan, liu2016makeup}.

Guo et al. \cite{guo2009digital} transferred face makeup from one image to another by decomposing face images into different layers, which were face structure, skin details, and color. Then, each layer of the example and original images were combined to obtain a natural makeup effect through gradient editing, weighted addition and alpha blending. Similarly, \cite{oo2016digital} decomposed the reference and target images into large scale layer, detail layer and color layer, where the makeup highlight and color information were transferred by Poisson editing, weighted means and alpha blending. \cite{lee2016anew} presented a method of makeup applying for different regions based on multiple makeup examples, e.g. the makeup of eyes and lip were taken from two references respectively.

\begin{figure}[htbp]
\centerline{\includegraphics[width=0.42\textwidth]{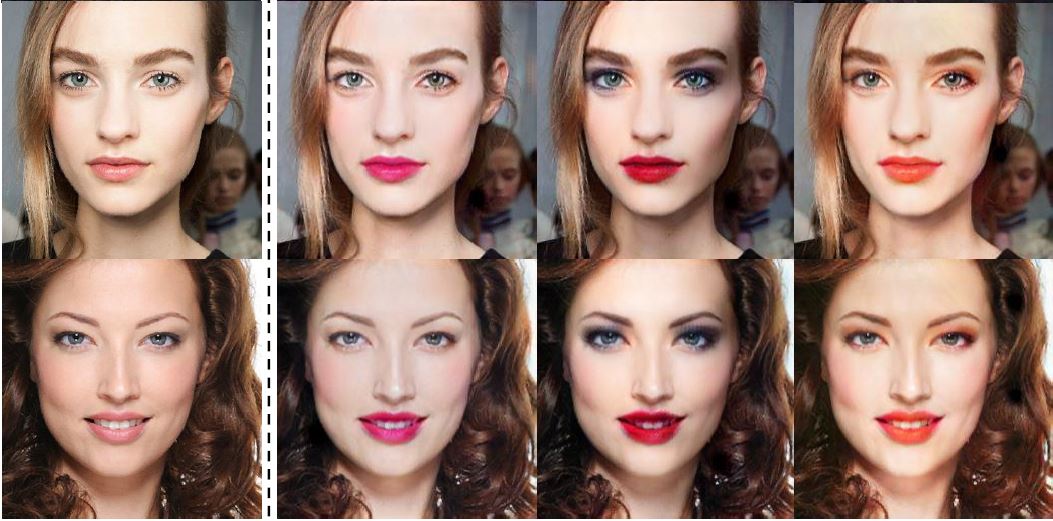}}
\caption{Examples of makeup transfer \cite{li2018beautygan}. The left column is the face images before makeup, and the remaining three columns are the makeup transfer result, where three makeup styles are translated.}
\label{MakeTrans}
\end{figure}

Traditional methods usually consider the makeup style as a combination of different components, and the overall output image usually looks unnatural with artifacts at adjacent regions \cite{li2018beautygan}. In contrast, the end-to-end deep networks acting on entire image showed great advantage in terms of output quality and diversity. For example, Liu et al. \cite{liu2016makeup} proposed a deep localized makeup transfer network to automatically generate faces with makeup by applying different cosmetics to corresponding facial regions in different manners. Alashkar et al. \cite{alashkar2017examples} presented an automatic makeup recommendation and synthesis system based on a deep neural network trained from pairwise of Before-After makeup images united with artist knowledge rules. Nguyen et al. \cite{nguyen2017smart} also proposed an automatic and personalized facial makeup recommendation and synthesis system. However, their system was realized based on a latent SVM model describing the relations among facial features, facial attributes and makeup attributes. With the rapid development of generative adversarial networks, they have been widly used in makeup transfer. \cite{chang2018pairedcyclegan} used cycle-consistent generative adversarial networks to simultaneously learn a makeup transfer function and a makeup removal function, such that the output of the network after a cycle should be consistent with the original input and no paired training data was needed. Similarly, BeautyGAN~\cite{li2018beautygan} applied a dual generative adversarial network. However, they further improved it by incorporating global domain-level loss and local instance-level loss to ensure an appropriate style transfer for both the global (identity and background) and independent local regions (like eyes and lip). Some transfer results of BeautyGAN are shown in \ref{MakeTrans}.

\subsection{Accessory Removal and Wearing}\label{sec:transtype:accessory}

The removal and wearing of accessories include those of the glasses, earrings, nose ring, nose stud, lip ring, etc. Among all the accessories, glasses are mostly common seen, as they are worn for various purposes including vision correction, bright sunlight prevention, eye protection, beauty, etc. Glasses could significantly affect the accuracy of face recognition as they usually cover a large area of human faces.

\begin{figure}[htbp]
\centerline{\includegraphics[width=0.45\textwidth]{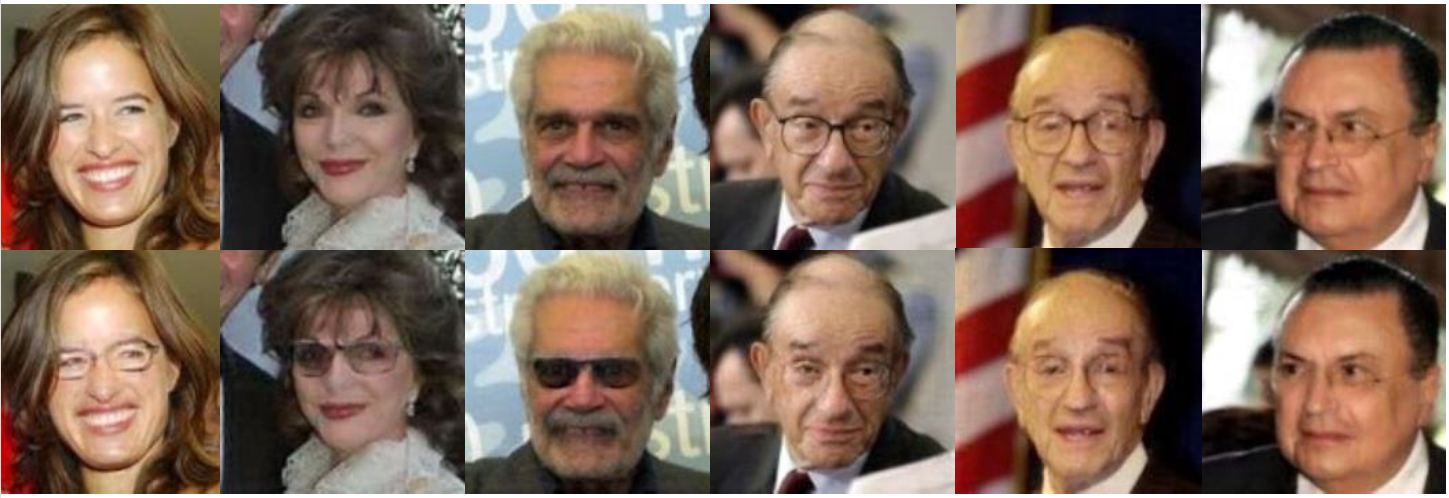}}
\caption{Examples of glasses wearing (the left three columns) and removal (the right three columns) \cite{shen2017learning}. The top row is the source images, and the bottom row is the transformation result.}
\label{GlassesTrans}
\end{figure}

Lv et al. \cite{lv2017data} synthesized glasses-wearing images using template-based method. Guo et al. \cite{guo2018face} fused virtual eyeglasses with face images by Augmented Reality technique. The InfoGAN proposed in \cite{chen2016infogan} learned disentangled representations of faces in a completely unsupervised manner, and was able to modify the presence of glasses. Shen et al. \cite{shen2017learning}proposed a face attribute manipulation method based on residual image, which was defined as the difference between the input image and the desired output image. They adopted two inverse image transformation networks to produce residual images for glasses wearing and removing. Some experiment results of \cite{shen2017learning} are shown in Fig. \ref{GlassesTrans}.

\subsection{Pose Transformation}\label{sec:transtype:pose}

The large pose discrepancy of head in the wild proposes a big challenge in face detection and recognition tasks, as self-occlusion and texture variation usually occur when head pose changes. Therefore, a variety of pose-invarient methods were proposed, including data augmentation for different poses. Since existing datasets mainly consist of near-frontal faces, which contradicts the condition for unconstrained face recognition.

\cite{feng2017dynamic} proposed a 2D profile face generator, which produced out-of-plane pose variations based on a PCA-based 2D shape model. Meanwhile many works used 3D face models for face pose translation \cite{zhu2016face, guo2017facenet3d, masi2016we, crispell2017dataset, lv2017data, kulkarni2015deep}. Facial texture is an important component of 3D face model, and directly affects the reality of generated faces. Deng et al. \cite{deng2018uv} proposed a Generative Adversarial Network (UV-GAN) to complete the facial UV map and recover the self-occluded regions. In their experiment, virtual instances under arbitrary poses were generated by attaching the completed UV map to the fitted 3D face mesh. In another way, Zhao et al.~\cite{zhao2017dual, zhao20183d} used network to enhance the realism of synthetic face images generated by 3D Morphable Model.

Generative models are pervasively employed by recent works to synthesize faces with arbitrary poses. \cite{van2016conditional} applied a conditional PixelCNN architecture to generate new portraits with different poses conditioned on pose embeddings. \cite{wiles2018x2face} introduced X2Face that could control a source face by a driving frame to produce a generated face with the identity of the source but the pose and expression of the other. Hu et al.~\cite{hu2018pose} proposed Couple-Agent Pose-Guided Generative Adversarial Network (CAPG-GAN) to realize flexible face rotation of arbitrary head poses from a single image in 2D space. They employed facial landmark heatmap to encode the head pose to control the generator. Moniz et al. \cite{moniz2018unsupervised} presented DepthNets to infer the depth of facial keypoints from the input image and predict 3D affine transformations that maps the input face to a desired pose and facial geometry. Cao et al. \cite{cao2018load} introduced Load Balanced Generative Adversarial Networks (LB-GAN), and decomposed the face rotation problem into two subtasks, which frontalize the face images first and rotate the front-facing faces later. Through their comparison experiment, LB-GAN performed better in identity preserving.

A special case of pose transformation is face frontalization. It is commonly used to increase the accuracy rate of face recognition by rotating faces to the front view, which are more friendly to the recognition model. Hassner et al. \cite{hassner2015effective} rotated a single, unmodified 3D reference face model for face frontalization. Taigman et al. \cite{taigman2014deepface} warped facial crops to frontal mode for accurate face alignment. TP-GAN \cite{huang2017beyond} and FF-GAN \cite{yin2017towards} are two classic face frontalization methods based on GANs. TP-GAN is a Two-Pathway Generative Adversarial Network for photorealistic frontal view synthesis by simultaneously perceiving global structures and local details. FF-GAN incorporates 3DMM into the GAN structure to provide shape and appearance priors for fast convergence with less training data. Furthermore, Zhao et al. \cite{zhao2018towards} proposed a Pose Invariant Model (PIM) by jointly learning face frontalization and facial features end-to-end for pose-invariant face recognition.

\begin{figure}[htbp]
\centerline{\includegraphics[scale=0.35]{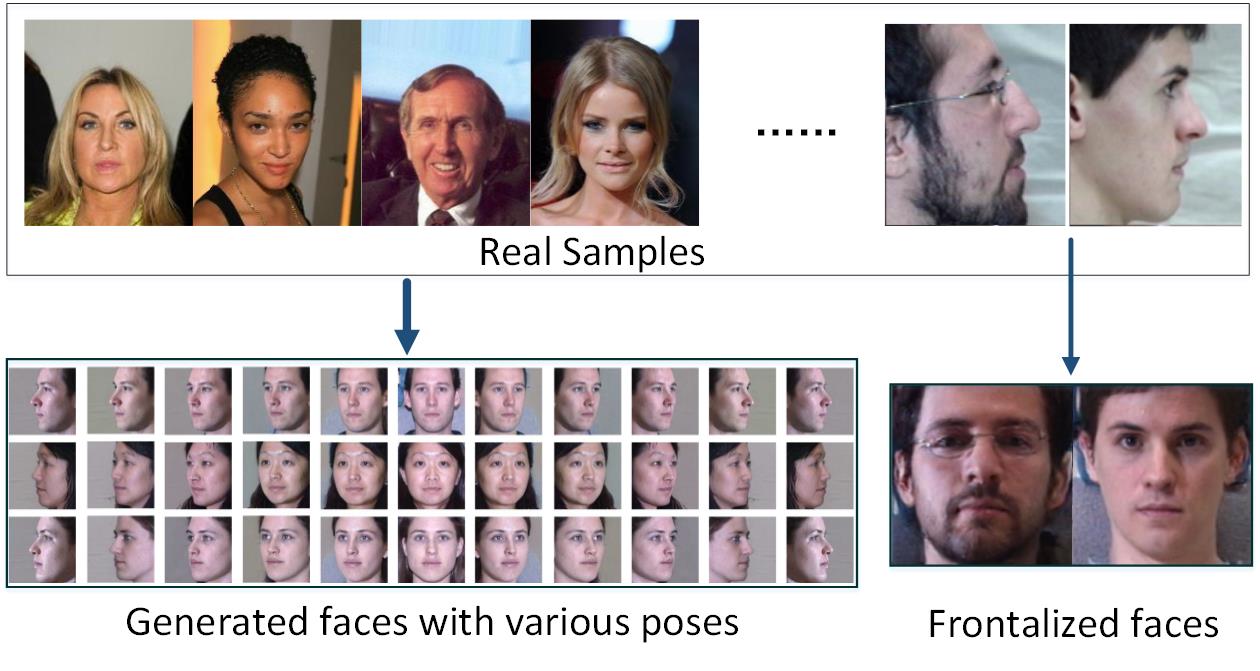}}
\caption{Examples of pose variation and frontalization. The real samples on the top left part are from CelebA \cite{liu2015deep}. The pose variation examples on the bottom left part are from \cite{hu2018pose}, and the frontalization examples on the right part are from \cite{huang2017beyond}.}
\label{PoseTrans}
\end{figure}

An illustration of face pose transformation is shown in Fig. \ref{PoseTrans}.

\subsection{Expression Synthesis and Transfer}\label{sec:transtype:expression}

\begin{figure}[htbp]
\centerline{\includegraphics[scale=0.45]{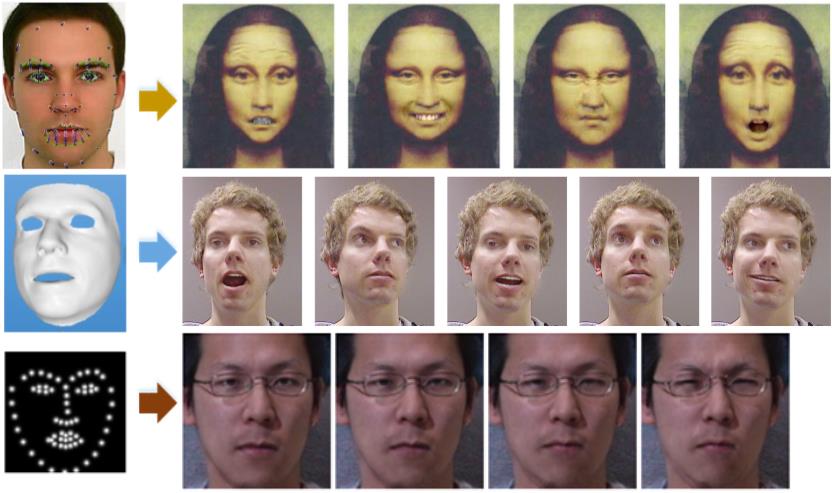}}
\caption{Some facial expression synthesis examples using 2D-based, 3D-based, and learning-based approaches. From top to bottom, the images are extracted from \cite{xie2018facial}, \cite{thies2015real}, and \cite{song2018geometry} respectively. The leftmost images illustrate mesh deformation, modified 3D face model, and input heatmap respectively.}
\label{ExpTrans}
\end{figure}

The facial expression synthesis and transfer technique is used to enrich the expressions (happy, sad, angry, fear, surprise, and disgust, etc.) of a given face and helps to improve the performance of tasks like emotion classification, expression recognition, and expression-invariant face recognition. The expression synthesis and transfer methods can be classified into 2D geometry based approach, 3D geometry based approach, and learning based approach (see Fig.~\ref{ExpTrans}).

The 2D and 3D based algorithms emerged earlier than learning-based methods, whose evident advantage is that they do not need a large amount of training samples. The 2D based algorithms transfer expressions relying on the geometry and texture features of the expressions in 2D space, such as \cite{xie2018facial}, while the 3D based algorithms generate face images with various emotions from 3D face models or 3D face data. \cite{agianpuye20133d} gave a comprehensive survey on the field of 3D facial expression synthesis. Until nowadays, "Blendshapes" model which was introduced in computer graphics remains the most prevalent approach for 3D expression synthesis and transfer \cite{masi2016we, thies2015real, guo2017facenet3d, kim2017deep}.

In recent years, a lot of learning based methods were proposed for expression synthesis and transfer. Besides the use of CNN~\cite{li2016deep, van2016conditional, flynn2016generating}, wide application of generative models of autoencoders and GANs has begun. For example, Yeh et al. \cite{yeh2016semantic} combined the flow-based face manipulation with variational autoencoders to encode the flow from one expression to another over a low-dimensional latent space. Zhou et al. \cite{zhou2017photorealistic} proposed the conditional difference adversarial autoencoder (CDAAE), which can generate specific expression for unseen person with a target emotion or facial action unit (AU) label. On the other side, the GAN-based methods can be further divided into four categories according to the generative condition of expression generation: reference images \cite{zhu2018emotion, bao2018towards}, emotion or expression codes~\cite{zhang2018joint, ding2018exprgan}, action unit labels~\cite{pham2018generative, pumarola2018ganimation}, and geometry~\cite{song2018geometry, qiao2018geometry, wu2018reenactgan}.																						

\cite{zhu2018emotion} and \cite{bao2018towards} generated different emotions with reference and target input. Zhang et al. \cite{zhang2018joint} generated different expressions under arbitrary poses conditioned by the expression and pose one-hot codes. Ding et al. \cite{ding2018exprgan} proposed an Expression Generative Adversarial Network (ExprGAN) for photo-realistic facial expression editing with controllable expression intensity. They designed an expression controller module to generate expression code, which was a real-valued vector containing the expression intensity description. Pham et al. \cite{pham2018generative} proposed a weakly supervised adversarial learning framework for automatic facial expression synthesis based on continuous action unit coefficients. Pumarola et al. \cite{pumarola2018ganimation} also controlled the generated expression by AU labels, and allowed a continuous expression transformation. In addition, they introduced an attention-based generator to promote the robustness of their model for distracting backgrounds and illuminations. Song et al. \cite{song2018geometry} proposed a Geometry-Guided Generative Adversarial Network (G2-GAN) to synthesize photo-realistic and identity-preserving facial images in different expressions from a single image. They employed facial geometry (fiducial points) as the controllable condition to guide facial texture synthesis. Qiao et al. \cite{qiao2018geometry} used geometry (facial landmarks) to control the expression synthesis with a facial geometry embedding network, and proposed a Geometry-Contrastive Generative Adversarial Network (GC-GAN) to transfer continuous emotions across different subjects even there were big shape difference. Furthermore, Wu et al. \cite{wu2018reenactgan} proposed a boundary latent space and boundary transformer. They mapped the source face into the boundary latent space, and transformed the source face's boundary to the target's boundary, which was the medium to capture facial geometric variances during expression transfer.

\subsection{Age Progression and Regression}\label{sec:transtype:age}

Age progression or face aging predicts one's future looks based on his current face, while age regression or rejuvenation estimates one's previous looks. All of them aim to synthesize faces of various ages and preserve personalized features at the same time. The generated face images enrich the data of individual subjects over a long range of age span, which enhances the robustness of the learned model to age variation.

The traditional methods of age transfer include the prototype-based method and model-based method. The prototype based method creates average faces for different age groups, learns the shape and texture transformation between these groups, and applies them to images for age transfer, such as \cite{kemelmacher2014illumination}. However, personalized features on individual faces are usually lost in this method. The model based method constructs parametric models of biological facial change with age, e.g. muscle, wrinkle, skin, etc. But such models typically suffer from high complexity and computational cost \cite{zhang2017age}. In order to avoid the drawbacks of the traditional methods, Suo et al. \cite{suo2010compositional} presented a compositional dynamic model, and applied a three level And-Or graph to represent the decomposition and the diversity of faces by a series of face component dictionaries. Similarly, \cite{shu2015personalized} learned a set of age-group specific dictionaries, and used a linear combination to express the aging process. In order to preserve the personalized facial characteristics, every face was decomposed into an aging layer and a personalized layer for consideration of both the general aging characteristics and the personalized facial characteristics.

\begin{figure}[htbp]
\centerline{\includegraphics[width=0.42\textwidth]{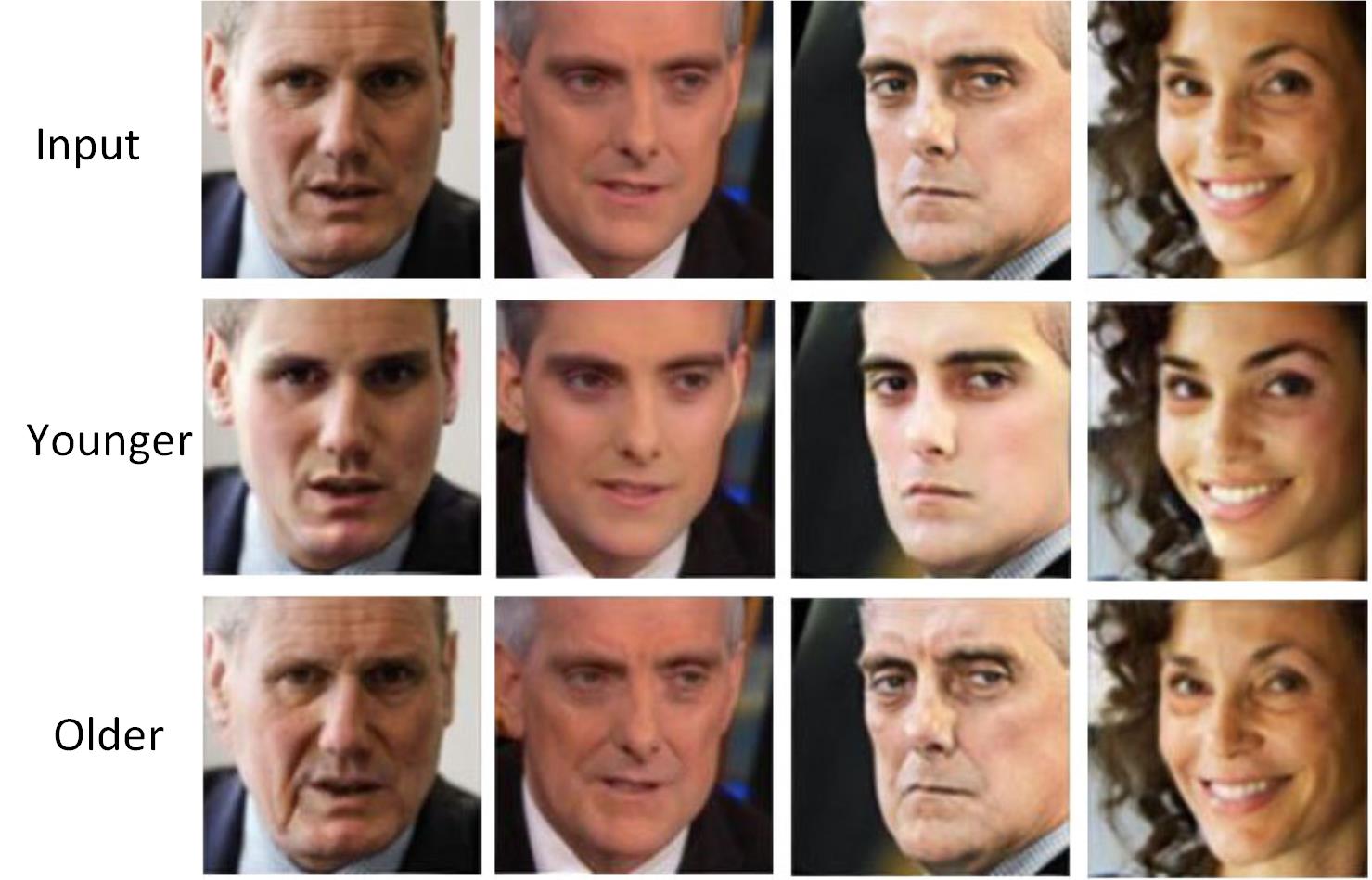}}
\caption{Examples of face rejuvenation and aging from \cite{zhao2018look}.}
\label{AgeTrans}
\end{figure}

More recent works applied GANs with encoders for age transfer. The input images are encoded into latent vectors, transformed in the latent space, and reconstructed back into images with a different age. Palsson et al. \cite{palsson2018generative} proposed three aging transfer models based on CycleGAN. Wang et al. \cite{wang2016recurrent} proposed a recurrent face aging (RFA) framework based on recurrent neural network. Zhang et al. \cite{zhang2017age} proposed a conditional adversarial autoencoder (CAAE) for face age progression and regression, based on the assumption that the face images lay on a high-dimensional manifold, and the age transformation could be achieved by a traversing along a certain direction. Antipov et al. \cite{antipov2017face} proposed Age-cGAN (Age Conditional Generative Adversarial Network) for automatic face aging and rejuvenation, which emphasized on identity preserving and introduced an approach for the optimization of latent vectors. Wang et al. \cite{wang2018face} proposed an Identity-Preserved Conditional Generative Adversarial Networks (IPCGANs). They combined a Conditional Generative Adversarial Network with an identity-preserved module and an age classifier to generate photorealistic faces in a target age. Zhao et al. \cite{zhao2018look} proposed a GAN-like Face Synthesis sub-Net (FSN) to learn a synthesis function that can achieve both face rejuvenation and aging with remarkable photorealistic and identity-preserving properties without the requirement of paired data and the true age of testing samples. Some results are shown in Fig. \ref{AgeTrans}. Zhu et al. \cite{zhu2018facial} paied more attention to the aging accuracy and utilized an age estimation technique to control the generated face. Li et al. \cite{li2018global} introduced a Wavelet-domain Global and Local Consistent Age Generative Adversarial Network (WaveletGLCA-GAN) for age progression and regression. In \cite{liu2018attribute}, Liu et al. pointed out that only identity preservation was not enough, especially for those models trained on unpaired face aging data. They proposed an attribute-aware face aging model with wavelet-based GANs to ensure attribute consistency.

\subsection{Other Styles Transfer}\label{sec:transtype:other}

\begin{figure}
\centerline{\includegraphics[width=0.43\textwidth]{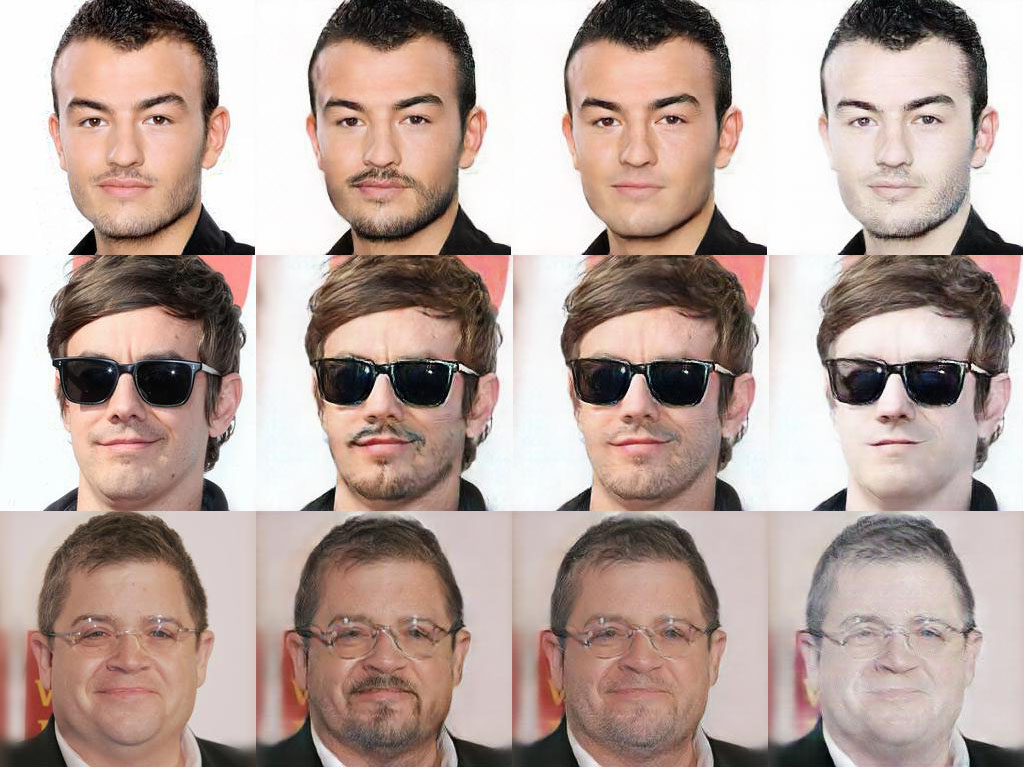}}
\caption{Face transformation examples created by AttGAN \citep{he2017attgan}. The leftmost column is the input (extracted from CelebA dataset\citep{liu2015deep}). From left to right, the remaining columns are mustache transfer result, beard transfer result, and skin color transfer result.}
\label{Others}
\end{figure}

In addition to the transformations summarized above, there are also some other types of transformations to enrich the face dataset, such as face illumination transfer \cite{leng2017data, zhuang2013single, bao2018towards, lv2017data, guo2017facenet3d, van2016conditional, crispell2017dataset, kulkarni2015deep}, gender transfer \cite{kim2017learning, choi2018stargan, li2016deep, shen2017learning}, skin color transfer \cite{choi2018stargan, he2017attgan}, eye color transfer \cite{chen2016infogan}, eyebrows transfer \cite{he2017attgan, lample2017fader}, mustache or beard transfer \cite{he2017attgan, kemelmacher2016transfiguring}, facial landmark perturbation \cite{lv2016landmark}, context and background change \cite{banerjee2018hallucinating}. Fig. \ref{Others} presents some examples of these transfer result.

In recent years, there have been more and more works trying to achieve multiple types of transformations through a unified neural network. For example, StarGAN~\cite{choi2018stargan} is able to perform expression, gender, age, and skin color transformations with a unified model and a one-time training. Conditional PixelCNN~\cite{van2016conditional} can generate images conditioned on expression, pose and illumination. Furthermore, the works~\cite{kim2017unsupervised, he2017attgan, lample2017fader} transferred or swapped multiple attributes (hair color, bang, pose, gender, mouth open, etc.) among different faces simultaneously. Recently, Sanchez et al. \cite{sanchez2018triple} proposed a triple consistency loss to bridge the gap between the distributions of the input and generated images, and allowed the generated images to be re-introduced to the network as input.

\section{Transformation Methods}\label{sec:methods}

In this section, we focus on the mapping function $\phi$ in Eq.~\ref{eq:map} by reviewing the existing methods on face data augmentation. We divide these methods into basic image processing, model-based transformation, realism enhancement, generative-based transformation, augmented reality, and auto augmentation (see Table~\ref{tab:method}), and give a clear description of them respectively.

\begin{table}[htbp]
\caption{An overview of transformation Methods}
\begin{center}
\begin{tabular}{m{2.3cm}<{\centering}|m{5.5cm}}
\Xhline{0.8pt}
\textbf{Methods}&\vspace{1mm}\hspace{10mm}\textbf{Implementation Examples}\vspace{1mm} \\
\Xhline{0.8pt}
\hspace{4mm}Basic\newline Image Processing&\vspace{1mm}noise adding, cropping, flipping, in-plane rotation, deformation, landmark perturbation, template fusion, mask blending, etc. \vspace{1mm} \\
\hline
\multirow{2}*{\begin{minipage}{0.1\textwidth}
\centering Model-based\newline Transformation
\end{minipage}}
&\vspace{1mm}2D models (e.g. 2D active appearance models) \\
&3D models (e.g. 3D morphable models)\vspace{1mm} \\
\hline
Realism Enhancement&\vspace{1mm}displacement map, augmentation function, detail refinement, domain adaption, etc.\vspace{1mm} \\
\hline
\vspace{1mm}Generative-based Transformation\vspace{1mm}&GANs, VAEs, PixelCNN, Glow, etc. \\
\hline
Augmented Reality&\vspace{1mm}real and virtual fusion\vspace{1mm} \\
\hline
Auto Augmentation&\vspace{1mm}neural net, search algorithm, etc.\vspace{1mm} \\
\Xhline{0.8pt}
\end{tabular}
\label{tab:method}
\end{center}
\end{table}

\subsection{Basic Image Processing}\label{sec:transmethod:improcess}

The geometric and photometric transformations for generic data augmentation mainly utilize the traditional image processing algorithms. Digital image processing is an important research direction in the field of computer vision, which contains many simple and complex algorithms with a wide range of applications, such as classification, feature extraction, pattern recognition, etc. Digital image transformation is a basic task of image processing, which can be expressed as:
\begin{equation}
g(x,y)=T[f(x,y)],
\end{equation}

in which $f(x,y)$ and $g(x,y)$ represent the input and output images respectively, and $T$ is the transformation function.

If $T$ is an affine transformation, it could realize image translation, rotation, scaling, reflection, shearing, etc. Affine transformation is an important class of linear 2D geometric transformations which maps pixel intensity value located at position $(x_1,y_1)$ in an input image to a new position $(x_2,y_2)$ in an output image. The affine transformation is usually written in homogeneous coordinates as:
\begin{equation}
\left[
\begin{array}{c}
x_2\\ y_2\\ 1\\
\end{array}
\right]
=
\left[
\begin{array}{ccc}
a_1&a_2&t_x \\
a_3&a_4&t_y \\
0&0&1 \\
\end{array}
\right]
\left[
\begin{array}{c}
x_1\\ y_1\\ 1\\
\end{array}
\right],
\end{equation}

where $(t_x,t_y)$ represents translation, and parameters ${a_i}$ contains rotation, scaling and shearing transformations.

If the image transformation is accomplished with a convolution between a kernel and an image, it can be used for image blurring, sharpening, embossing, edge enhancement, noise adding, etc. An image kernel $k$ is a small two-dimensional matrix, and the image convolution can be expressed by the following equation:
\begin{flalign}
\begin{split}
g(x,y)&=k(x,y){\ast}f(x,y) \\
&=\sum_{s=-a}^a \sum_{t=-b}^b k(s,t)f(x-s,y-t).
\end{split}
\end{flalign}

Meanwhile, the image color can also be linearly transformed through:
\begin{equation}
g(x,y)=w(x,y){\cdot}f(x,y)+b,
\end{equation}

where $w(x,y)$ is the weight and \(b\) is the color value bias. This transformation can be performed in various color spaces including RGB, YUV, HSV, etc.

Some works performed a sequence of image manipulations to implement complex transformation for face data augmentation. For example,~\cite{lv2017data} perturbed the facial landmarks through a series of operations, including facial landmarks location, perturbation of the landmarks by Gaussian distribution and image normalization.~\cite{xie2018facial} used face deformation and wrinkle mapping for facial expression synthesis. ~\cite{hu2018frankenstein} proposed a pure 2D method to generate synthetic face images by compositing different face parts (eyes, nose, mouth) of two subjects. ~\cite{banerjee2017srefi} synthesized new faces by stitching region-specific triangles from different images, which were proximal in a CNN feature representation space.

\subsection{Model-based Transformation}\label{sec:transmethod:model}

Model-based face data augmentation fits a face model to the input image and synthesizes faces with different appearance by varying the parameters of the fitted model. The commonly used generative face models can be classified as 2D and 3D, and the most representative models are 2D Active Appearance Models (2D AAMs)~\cite{cootes2001active} and 3D Morphable Models (3DMMs)~\cite{blanz1999morphable}. Both the AAMs and 3DMMs consist of a linear shape model and a linear texture model. The main difference between them is the shape component, which is 2D for AAM and 3D for 3DMM.

AAM is a parameterized face model represented as the variability of the face shape and texture. It is constructed based on a representative training set through a statistical based template matching method, and computed using Principal Component Analysis (PCA). The shape of an AAM is described as a vector of coordinates from a series of landmark points, and a landmark connectivity scheme. Mathematically, a shape of a 2D AAM is represented by concatenating $n$ landmark points ${(x_i,y_i)}$ into a vector $(x_1,y_1,x_2,y_2,{\ldots},x_n,y_n)^T$.

With all the shape vectors from the training set, a mean shape is extracted. Then, the PCA is applied to search for directions that have large variance in the shape space and project the shape vectors onto it. Finally, the shape is modeled as a base shape plus a linear combination of shape variations:

\begin{equation}
{\bf s}=\overline{\bf s} + {\bf P}_s{\cdot}{\bf b}_s.
\label{eq:aamshape}
\end{equation}

In Eq.~\ref{eq:aamshape}, $\overline{\bf s}$ denotes the base shape or mean shape, ${\bf P}_s$ is a set of orthogonal modes of variation obtained from the PCA eigenvectors. The coefficients ${\bf b}_s$ include the shape parameters in the shape subspace.

\begin{figure}[htbp]
\centerline{\includegraphics[scale=0.4]{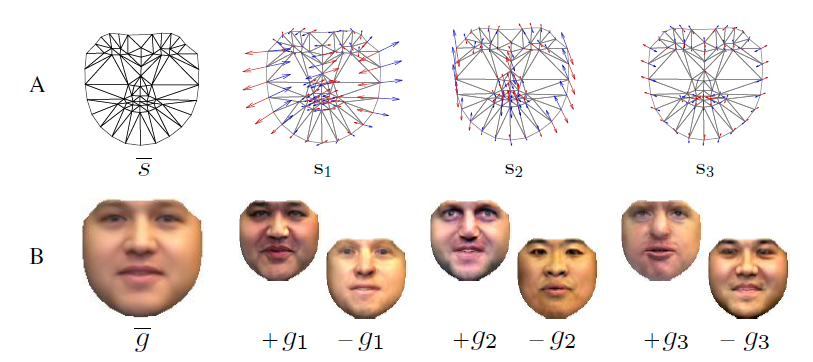}}
\caption{2D AAM illustration \cite{matthews20072d}. (A)The 2D AAM shape variation. (B)The 2D AAM texture variation.}
\label{AAMFig}
\end{figure}

In AAM, the texture is defined with the pixel intensities. For $t$ pixels sampled, the texture is expressed as ${[g_1,g_2,{\ldots},g_{ct}]}^T$, where $c$ is the number of channels in each pixel. After a piece-wise affine warping and photometric normalization for all the texture vectors extracted from the training set, the texture model is obtained using PCA and has a similar expression with the shape model:

\begin{equation}
{\bf g}=\overline{\bf g} + {\bf P}_g{\cdot}{\bf b}_g,
\label{eq:aamtex}
\end{equation}

where $\overline{\bf g}$ is the mean texture, ${\bf P}_g$ is the matrix consisting of a set of orthonormal base vectors, and ${\bf b}_g$ includes the texture parameters in the texture subspace. Fig.~\ref{AAMFig} shows an example of 2D AAM constructed from 5 people using approximately 20 training images for each person. Fig.~\ref{AAMFig}A shows the mean shape $\overline{\bf s}$ and the first three shape variation components ${\bf s}_1$, ${\bf s}_2$ and ${\bf s}_3$. Fig.~\ref{AAMFig}B shows the mean texture $\overline{\bf g}$ and an illustration of the texture variation, where $+{\bf g}_j$ and $-{\bf g}_j$ denote the addition and subtraction of the $j$th texture mode to the mean texture respectively.

\begin{figure}[htbp]
\centerline{\includegraphics[scale=0.32]{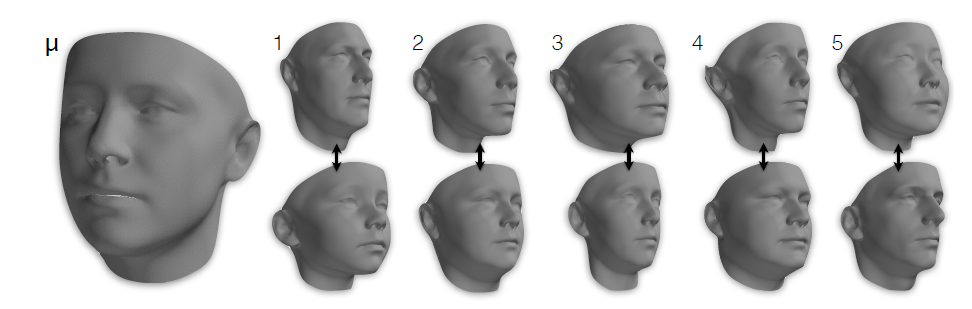}}
\caption{Visualization of the shape of LSFM \cite{booth20163d}. The leftmost shows the visualization of the mean shape. The others are the visualizations of the first five principal components of shape, with each visualized as additions and subtractions from the mean.}
\label{DDMFig}
\end{figure}

3D Morphable Model is constructed from a set of 3D face scans with dense correspondence. The geometry of each scan is represented by a shape vector $\textbf{S}=(x_1,y_1,z_1,{\ldots},x_n,y_n,z_n)$, that contains the 3D coordinates of its $n$ vertices, and the texture of the scan is represented by a texture vector $\textbf{T}=(r_1,g_1,b_1,{\ldots},r_n,g_n,b_n)$, that contains the color values of the $n$ corresponding vertices. A morphable face model is constructed based on PCA and expressed as:

\begin{equation}
\begin{split}
{\bf S}_{model}=\overline{\bf S}+\sum\limits_{i = 1}^{m} {{\alpha}_i}{{\bf s}_i}, \\
{\bf T}_{model}=\overline{\bf T}+\sum\limits_{i = 1}^{m} {{\beta}_i}{{\bf t}_i}, \\
\end{split}
\end{equation}

where $m$ is the number of eigenvectors. $\overline{\bf S}$ and $\overline{\bf T}$ represent the mean shape and mean texture respectively. ${\bf s}_i$ and ${\bf t}_i$ are the $i$th eigenvectors, and $\alpha=({\alpha}_1,{\alpha}_2,{\ldots},{\alpha}_m)$ and $\beta=({\beta}_1,{\beta}_2,{\ldots},{\beta}_m)$ are shape and texture parameters respectively. 3DMM has very similar model expressions with 2D AAM. Fig. \ref{DDMFig} shows an example of 3DMM -- LSFM~\cite{booth20163d}, which was constructed from 9,663 distinct facial identities. In this figure, we can visualize the mean shape of LSFM model along with the top five principal components of the shape variation.

To generate model instances (images), AAMs use a 2D image normalization and a 2D similarity transformation, while the 3DMMs use the scaled orthographic model or weak perspective model. More detailed description about the representational power, constriction, and real-time fitting of the AAMs and 3DMMs can be found in \cite{matthews20072d}.

\begin{figure}[htbp]
\centerline{\includegraphics[scale=0.17]{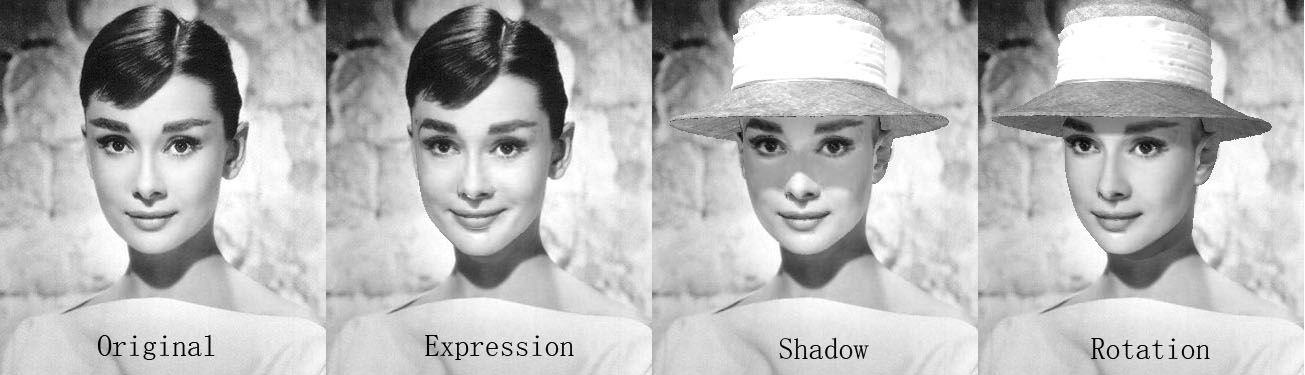}}
\caption{Face reconstruction based on 3DMM \cite{blanz1999morphable}. The left image is the original 2D image. After a 3D face reconstruction, new facial expression, shadow, and pose can be generated.}
\label{MMFig}
\end{figure}

One advantage of 3D model is the availability of surface normals, which can be used to simulate the lighting effect (Fig. \ref{MMFig}). Therefore, some works combined the above 3DMMs with illumination modelled by Phone model \cite{blanz2003face} or Spherical Harmonic model \cite{zhang2006face} to generate more realistic face images. \cite{hu2016face} classified the 3d face models into 3DSM (3D Shape Model), 3DMM and E-3DMM (Extended 3DMM). The 3DSM can only explicitly model pose. In contrast, 3DMM can model pose and illumination, while E-3DMM can model pose, illumination and facial expression. Furthermore, in order to overcome the limitation of linear models, Tran et al.~\cite{tran2018nonlinear} utilized neural networks to reconstruct nonlinear 3D face morphable model which is a more flexible representation. Hu et al.~\cite{hu2016face} proposed U-3DMM (Unified 3DMM) to model more intra-personal variations, such as occlusion.

Matching an AAM to an image can be considered as a registration problem and solved via energy optimization. In contrast, 3D model fitting is more complicated. The fitting is manly conducted by minimizing the color value differences over all the pixels in the facial region between the input images and its model-based reconstruction result. However, as the fitting is an ill-posed problem, it is not easy to get an efficient and accurate fitting~\cite{hu2016face}. \cite{masi2016we} used the corresponding landmarks from the 2D input image and the 3D model to calculate the extrinsic camera parameters. \cite{guo2017facenet3d} and \cite{thies2015real} applied the analysis-by-synthesis strategy to do model fitting. In recent years, many works use neural networks to do 3D face model fitting, such as \cite{zhu2016face} and \cite{guo2017facenet3d}. More fitting methods can be found in \cite{hu2016face}.

Although 3D model can be used to generate more diverse and accurate transformations of faces, there still exist some challenges. One of the challenges is the visualization of teeth and mouth cavity, as the face model only represents the skin surface and dose not include eyes, teeth, and mouth cavity. \cite{thies2015real} used two textured 3D proxies for the teeth simulation, and achevied mouth transformation by warpping a static frame of an open mouth based on the tracked landmarks. Another challenge is the artifacts caused by the missing of occluded regions when the head pose is changed. Zhu et al. \cite{zhu2015high} proposed an inpainting method which made use of Possion editing to estimate the mean face texture and fill the facial detail of the invisible region caused by self-occlusion.

\subsection{Realism Enhancement}\label{sec:transmethod:real}

As illustrated in Fig.~\ref{AugFig}, besides the direct style transfer from real 2D images, making simulated samples more realistic is an important method of face data augmentation. Although modern computer graphics techniques provide powerful tools to generate virtual faces, it still remains difficult to generate a large number of photorealistic samples due to the lack of accurate illumination and complicated surface modeling. The state-of-the-art synthesizes virtual faces using a morphable model and has difficulty in generating detailed photorealistic images of faces, such as faces with wrinkles~\cite{guo2017facenet3d}. Anyway, the simulation process involved is simplified for the consideration of the speed of modeling and rendering, making the generated images not realistic enough. In order to improve the quality of simulated samples, some realism enhancement techniques were proposed.

\begin{figure}[htbp]
\centerline{\includegraphics[width=0.4\textwidth]{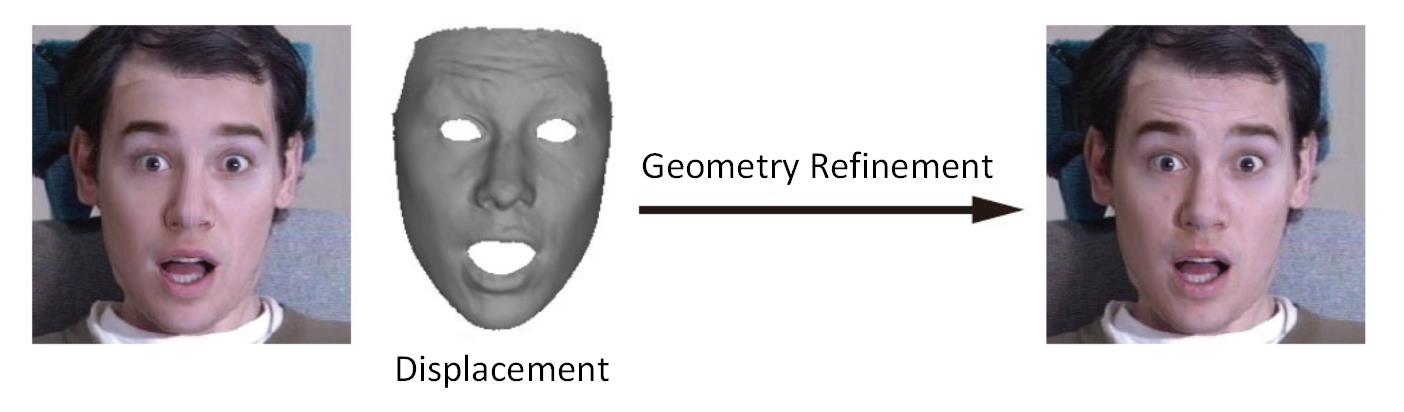}}
\caption{Geometry refinement by displacement map \cite{guo2017facenet3d}. The right image has more face details (e.g. wrinkles) than the left input.}
\label{DisMap}
\end{figure}

\begin{figure}[htbp]
\centerline{\includegraphics[width=0.5\textwidth]{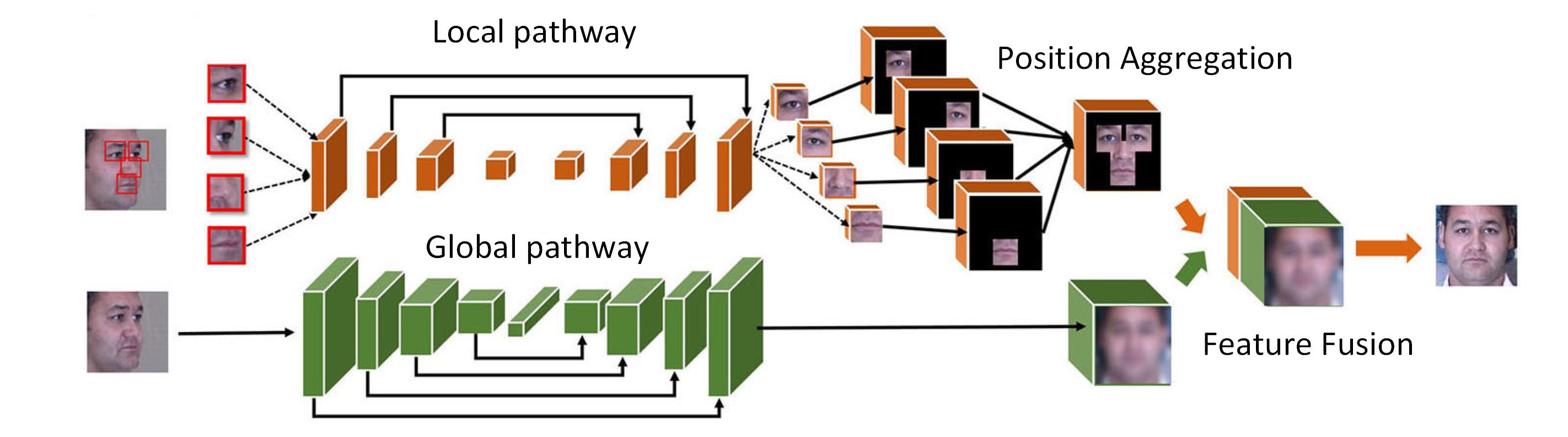}}
\caption{The global and local pathways of the generator in \cite{huang2017beyond}.}
\label{real-gl}
\end{figure}

Guo et al.~\cite{guo2017facenet3d} introduced displacement map that encoded the geometry details of face in a displacement along the depth direction of each pixel. It could be used for face detail transfer and synthesizing fine-detailed faces (Fig. \ref{DisMap}). In~\cite{zhao20183d}, Zhao et al. applied GANs to improve the realism of face simulator's output by making use of unlabeled real faces. Specifically, they fed the defective simulated faces obtained from a 3D morphable model into a generator for realism refinement, and used two discriminators to minimize the gap between real domain and virtual domain by discriminating real \emph{v.s.} fake and preserving identity information. Furthermore, Gecer et al.~\cite{gecer2018semi} introduced a two-way domain adaption framework similar to CycleGAN to improve the realism of rendered faces. \cite{zhao2018towards} and \cite{huang2017beyond} applied dual-path GANs, which contained separate global generator and local generators for global structure and local details generation (see Fig. \ref{real-gl} for illustration). Shrivastava et al.~\cite{shrivastava2017learning} proposed SimGAN, a simulated and unsupervised learning method to improve the realism of synthetic images using a refiner network and adversarial training. Sixt et al.~\cite{sixt2018rendergan} embedded a simple 3D model and a series of parameterized augmentation functions into the generator network of GAN, where the 3D model was used to produce virtual samples from input labels and the augmentation functions used to add the missing characteristics to the model output. Although the proposed realism enhancement methods in~\cite{shrivastava2017learning} and~\cite{sixt2018rendergan} were not originally aimed at face images, they have the potential to be applied to realistic face data generation.

\subsection{Generative-based Transformation}\label{sec:transmethod:generative}

\begin{figure}[htbp]
\centerline{\includegraphics[width=0.49\textwidth]{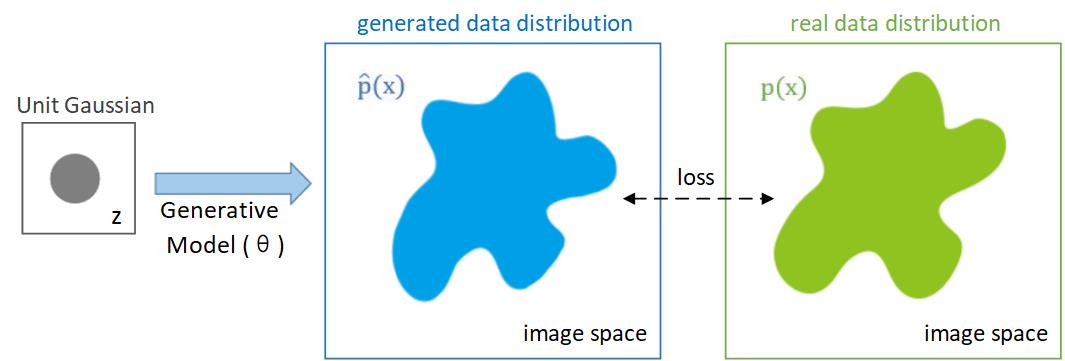}}
\caption{A schematic diagram of generative model. The unit Gaussian distribution is mapped to a generated data distribution by the generative model. And the distance between the generated data distribution and the real data distribution is measured by the loss.}
\label{GenFig}
\end{figure}

The generative models provide a powerful tool to generate new data from modeled distribution by learning the data distribution of the training set. Mathematically, the generative model can be expressed as follows. Suppose there is a dataset of examples $\{x_1,...,x_n\}$ as samples from a real data distribution $p(x)$ as illustrated in Fig.~\ref{GenFig}, in which the green region shows a subspace of the image space containing real images. The generative model maps a unit Gaussian distribution (grey) to another distribution $\hat{p}(x)$ (blue) through a neural network, which is a function with parameters $\boldsymbol{\theta}$. The generated distribution of images can be tweaked when the network parameters are changed. Then the aim of the training process is to start from random and find parameters $\boldsymbol{\theta}$ that produce a distribution that closely matches the real data distribution.

In recent years, the deep generative models have attracted much attention and significantly promoted the performance of data generation. Among them, the three most popular models are Autoregressive Models, Variational Autoencoders (VAEs), and Generative Adversarial Networks. Autoregressive Models and VAEs aim to minimize the Kullback-Liebler(KL) divergence between the modeled distribution and the real data distribution. In contrast, the Generative Adversarial Networks apply adversarial learning to generate data indistinguishable from the real samples, and hence avoid specifying an explicit density for any data point, which belong to the class of implicit generative models \cite{grover2018flow}.

\subsubsection{Autoregressive Generative Models}\label{sec:transmethod:generative:autoregressive}
The typical examples of autoregressive generative models are PixelRNN and PixelCNN proposed by Oord et al.~\cite{oord2016pixel}. They tractably model the joint distribution of the pixels in the image by decomposing it into a product of conditional distributions, which can be formulated as:

\begin{equation}
p({\bf x})=\prod_{i=1}^{n^2} p(x_i|x_1,...,x_{i-1}),
\label{eq:pixelcnn}
\end{equation}

in which $p(\textbf{x})$ is the probability of image $\textbf{x}$ formed of $n{\times}n$ pixels, and the value $p(x_i|x_1,...,x_{i-1})$ is the probability of the \emph{i}-th pixel $x_i$ given all the previous pixels $x_1,...,x_{i-1}$. Thus, the image modeling problem turns into a sequential problem, where one learns to predict the next pixel given all the previously generated pixels (Fig.~\ref{GenModel}-left). The PixelRNN models the pixel distribution with two-dimensional LSTM, and PixelCNN models with convolutional networks. Oord et al.~\cite{van2016conditional} further presented Gated PixelCNN and Conditional PixelCNN. The former replaced the activation unit in the original pixelCNN with gated block, and the latter modeled the complex conditional distributions of natural images by introducing conditional variant to the latent vector. In addition, Salimans et al. proposed PixelCNN++ \cite{salimans2017pixelcnn++} which simplified PixelCNN's structure and improved the synthetic images' quality.

\subsubsection{Variational Autoencoders}\label{sec:transmethod:generative:vae}

Variational Autoencoders formalize the data generation problem in the framework of probabilistic graphical models rooted in Bayesian inference. The idea of VAEs is to learn the latent variables, which are low-dimensional latent representations of the training data and inferred through a mathematical model. In Fig.~\ref{GenModel}-mid, the latent variables are denoted by $\textbf{z}$, and the probability distribution of $\textbf{z}$ is denoted as $p_{\boldsymbol{\theta}}(\textbf{z})$, where $\boldsymbol{\theta}$ are the model parameters. There are two components in a VAE: the encoder and the decoder. The encoder encodes the training data $\textbf{x}$  into a latent representation, and the decoder maps the obtained latent representation $\textbf{z}$ back to the data space. In order to maximize the likelihood of the training dataset, we maximize the probability $p_{\boldsymbol{\theta}}(\textbf{x})$ of each data:

\begin{equation}
p_{\boldsymbol{\theta}}({\bf x})=p_{\boldsymbol{\theta}}({\bf x},{\bf z}){\rm d}{\bf z}=p_{\boldsymbol{\theta}}({\bf x}|{\bf z})p_{\boldsymbol{\theta}}({\bf z}){\rm d}{\bf z}.
\label{eq:vae}
\end{equation}

The above integral is intractable, and the true posterior density $p_{\boldsymbol{\theta}}(\textbf{z}|\textbf{x})=p_{\boldsymbol{\theta}}(\textbf{x}|\textbf{z})p_{\boldsymbol{\theta}}(\textbf{z})/p_{\boldsymbol{\theta}}(\textbf{x})$ is intractable, so the EM (Expectation-Maximization) algorithm cannot be used. The required integrals for any reasonable mean-field VB(variational Bayesian) algorithm are also intractable \citep{kingma2013auto}. Therefore the VAEs turn to infer $p(\textbf{z}|\textbf{x})$ using variational inference which is a basic optimization problem in Bayesian statistics. They first model $p(\textbf{z}|\textbf{x})$ using simpler distribution $q_{\boldsymbol{\phi}}(\textbf{z}|\textbf{x})$ which is easy to find and try to minimize the difference between $p_{\boldsymbol{\theta}}(\textbf{z}|\textbf{x})$ and $q_{\boldsymbol{\phi}}(\textbf{z}|\textbf{x})$ using KL divergence metric approach. The marginal likelihood of individual datapoint can be written as:

\begin{equation}
\log{p_{\boldsymbol{\theta}}({\bf x})}=D_{KL}(q_{\boldsymbol{\phi}}({\bf z}|{\bf x})||p_{\boldsymbol{\theta}}({\bf z}|{\bf x}))+{\cal{L}}({\boldsymbol{\theta}},{\boldsymbol{\phi}};{\bf x}),
\label{eq:vae_kl}
\end{equation}

where the first term of the right-hand side is the KL divergence of the approximate from the true posterior. Since this divergence is non-negative, the second term ${\cal{L}}({\boldsymbol{\theta}},{\boldsymbol{\phi}};x)$ represents the lower bound on the marginal likelihood of the datapoint which we want to optimize and can be written as:

\begin{equation}
{\cal{L}}({\boldsymbol{\theta}},{\boldsymbol{\phi}};{\bf x})={\mathbb{E}}_{q_{\boldsymbol{\phi}}({\bf z}|{\bf x})}[\log{p_{\boldsymbol{\theta}}({\bf x}|{\bf z})}] \\
-D_{KL}(q_{\boldsymbol{\phi}}({\bf z}|{\bf x})||p_{\boldsymbol{\theta}}({\bf z})).
\label{eq:vae_L}
\end{equation}

More detailed explanation of the mathematical derivation can be found in \cite{kingma2013auto}.

In order to control the generation direction of the VAEs, Sohn et al.~\cite{sohn2015learning} proposed a conditional variational auto-encoder (CVAE), which is a conditional directed graphical model being trained to maximize the conditional log-likelihood. Pandey et al.~\cite{pandey2017variational} introduced conditional multimodel autoencoder (CMMA) to address the problem of conditional modality learning through the capture of conditional distribution. Their model can be used to generate and modify faces conditioned on facial attributes. Another approach to generate faces from visual attributes was introduced by \cite{yan2016attribute2image}. They modeled the image as a composition of foreground and background, and developed a layered generative model with disentangled latent variables that were learned using a variational auto-encoder. Huang et al.~\cite{huang2018introvae} proposed an introspective variational autoencoder (IntroVAE) model for synthesizing high-resolution photorealistic images by self-evaluating the quality of the generated samples during the training process. They borrowed the idea of GANs and reused the encoder as a discriminator to calssify the generated and training samples.

\subsubsection{Generative Adversarial Network}\label{sec:transmethod:generative:gan}

Generative Adversarial Network is an alternative framework to train generative models which gets rid of the difficult approximation of intractable probabilistic computations. It takes game-theoretic approach and plays an adversarial game between a generator and a discriminator. The discriminator learns to distinguish between real and fake samples, while the generator learns to produce fake samples that are indistinguishable from real samples by the discriminator.

\begin{figure}[htbp]
\centerline{\includegraphics[scale=0.21]{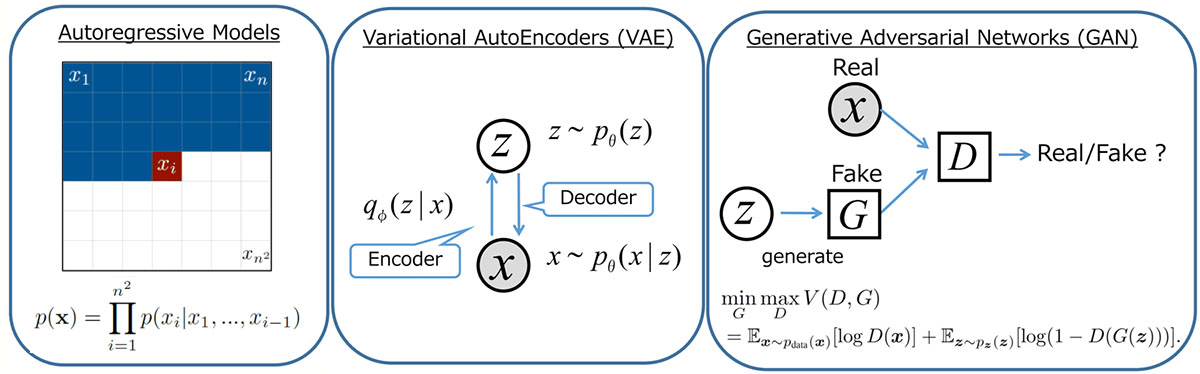}}
\caption{Comparison of generative models (extracted from STAT946F17 at the University of Waterloo \cite{statwiki2017conditional}).}
\label{GenModel}
\end{figure}

As shown in Fig.~\ref{GenModel}-right, in order to learn a generated distribution $p_g$ over data $\textbf{x}$, the generator builds a mapping function from a prior noise distribution $p_\textbf{z}(\textbf{z})$ to a data space as $G(\textbf{z};\boldsymbol{\theta}_g)$, where $G$ is a differentiable function represented by a multilayer perceptron with parameters $\boldsymbol{\theta}_g$. The discriminator, whose mapping function is denoted as $D(\textbf{x};\boldsymbol{\theta}_d)$, outputs a single scalar representing the probability that $\textbf{x}$ comes from the training data rather than $p_g$. $G$ and $D$ are trained simultaneously by adjusting the parameters of $G$ to minimize $\log{(1-D(G(\textbf{z})))}$ and also the parameters of $D$ to maximize the probability of assigning the correct label to both training examples and generated samples. The objective can be represented by the following two-player min-max game with value function $V(D,G)$:

\begin{flalign}
\begin{split}
\min \limits_{G} \max \limits_{D} V(D,G)=&{\mathbb{E}}_{{\bf x}{\thicksim}p_{data}({\bf x})}[\log{D({\bf x})}]+\\
&{\mathbb{E}}_{{\bf z}{\thicksim}p_{\bf z}({\bf z})}[\log{(1-D(G({\bf z})))}].
\end{split}
\end{flalign}

Ever since the proposal of the basic GAN concept, researchers have been trying to improve its stability and capability. Radford et al.~\cite{radford2015unsupervised} introduced DCGANs, the deep convolutional generative adversarial networks, to replace the multi-layer perceptrons in the basic GANs with convolutional nets. Whereafter, a lot of improvements for DCGANs were proposed. For example,~\cite{salimans2016improved} improved the training techniques to encourage convergence of the GANs game, ~\cite{arjovsky2017wasserstein} applied Wasserstein distance and improved the stability of learning, Conditional GANs could determine the specific representation of the generated images by feeding the condition to both the generator and discriminator~\cite{mirza2014conditional, isola2017image}. What's more, Zhang et al.~\cite{zhang2017stackgan} proposed a two-stage GANs--StackGAN. The Stage-I GAN sketched the primitive shape and colors of the object, and the Stage-II GAN generated realistic high-resolution images based on the Stage-I's output. Chen et al.~\cite{chen2016infogan} introduced InfoGAN, an information-theoretic extension to the basic GANs. It used a part of the input noise vector as latent code to target the salient structured semantic features of the data distribution in an unsupervised way.

\begin{figure}[htbp]
\centerline{\includegraphics[scale=0.26]{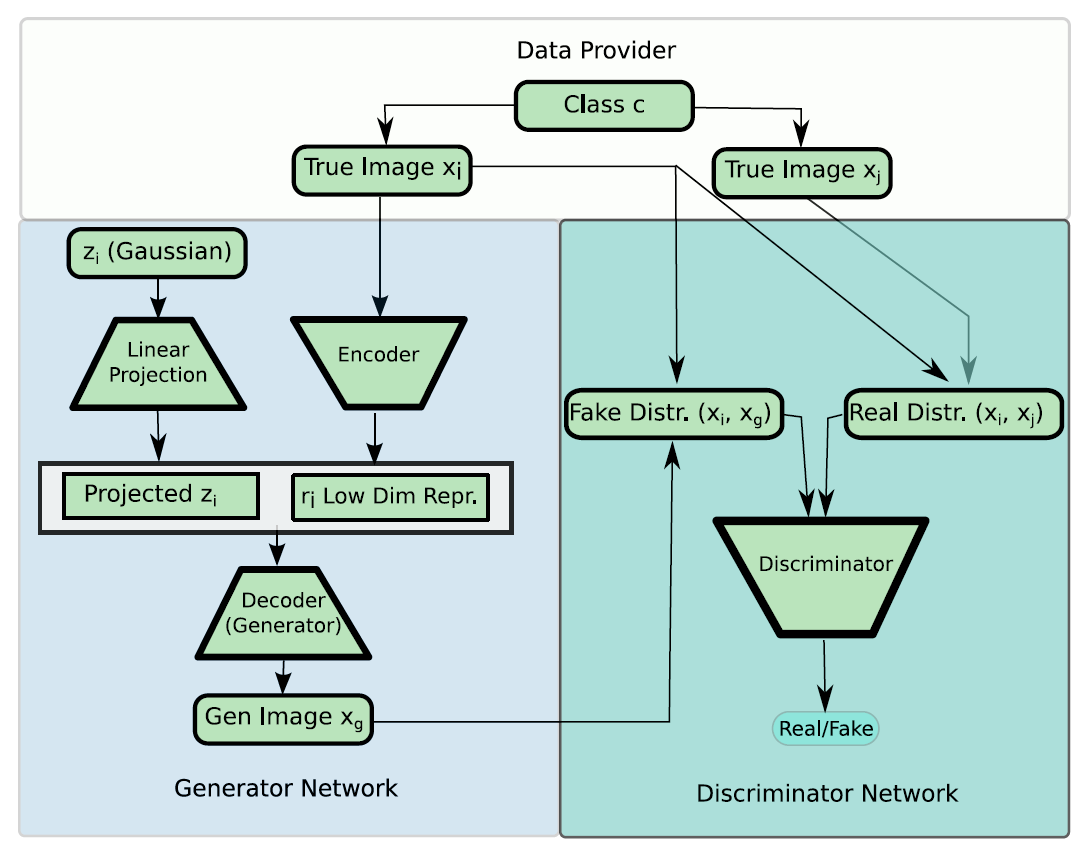}}
\caption{DAGAN Architecture \cite{antoniou2017data}. The DAGAN comprises of a generator network and a discriminator network. Left: During the generation process, an encoder maps the input image into a lower-dimensional latent space, while a random vector is transformed and concatenated with the latent vector. Then the long vector is passed to the decoder to generate an augmentation image. Right: In order to ensure the realism of the generated image, an adversarial discriminator network is employed to discriminate the generated images from the real images.}
\label{DAGAN}
\end{figure}

Since the first proposition of GANs~\cite{goodfellow2014generative} in 2014, it has attracted much attention because of its remarkable performance in a wide range of applications, including face data augmentation. Antoniou et al.~\cite{antoniou2017data} proposed Data Augmentation Generative Adversarial Network (DAGAN) based on conditional GAN (cGAN) and tested its effectiveness on vanilla classifiers and one shot learning. Fig.~\ref{DAGAN} shows the architecture of DAGAN which is a basic framework for data augmentation based on cGAN. Actually, many face data augmentation works followed this architecture and extended it to a more powerful network.

\begin{figure}[htbp]
\centerline{\includegraphics[scale=0.3]{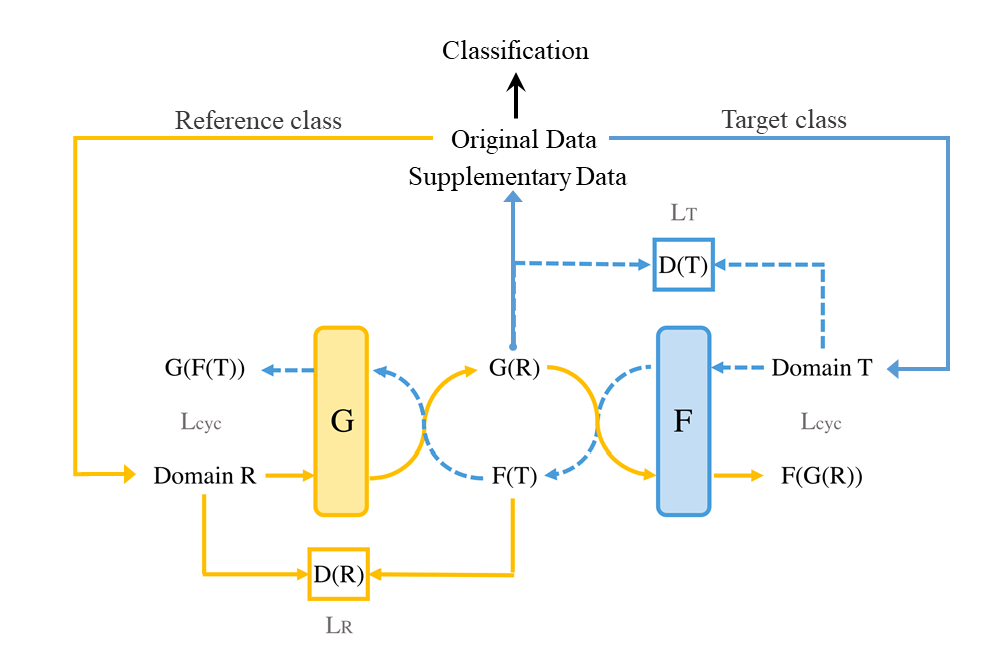}}
\caption{Framework proposed by \cite{zhu2018emotion}. The reference and target image domians are represented by $R$ and $T$ respectively. $G$ and $F$ are two generators to transfer $R$$\rightarrow$$T$ and $T$$\rightarrow$$R$. The discriminators are represented by $D(R)$ and $D(T)$ respectively, where $D(R)$ aims to distinguish between the real images in $R$ and the generated fake images in $F(T)$, and $D(T)$ aims to distinguish between the real images in $T$ and the generated fake images in $G(R)$. What's more, the cycle-consistency loss was used to guarantee that $F(G(R))$$\approx$$R$ and $G(F(T))$$\approx$$T$.}
\label{CycleGAN}
\end{figure}

Zhu et al.~\cite{zhu2018emotion} presented another basic framework for face data augmentation based on CycleGAN \cite{zhu2017unpaired}. Similar to cGAN, CycleGAN is also an general-purpose solution for image-to-image translation, but it learns a dual mapping between two domains simultaneously with no need for paired training examples, because it combines a cycle consistency loss with adversarial loss. \cite{zhu2018emotion} used this framework (whose architecture is shown in Fig.~\ref{CycleGAN}) to generate auxiliary data for unbalanced dataset, where the data class with fewer samples was selected as transfer target and the data class with more samples was reference. In \cite{zhu2018emotion}, the authors made a comparison between CycleGAN and the classical GAN. They claimed that the original GAN learns a mapping from low-dimensional manifold (determined by noise) to high-dimensional data spaces (images), while CycleGAN learns the translation between two high dimensional data domains. Thus, CycleGAN can complete and complement an imbalanced dataset more efficiently.

Except the above two basic frameworks of face data augmentation based on GAN, numerous extended approaches were proposed in recent years, such as DiscoGAN \cite{kim2017learning}, StarGAN \cite{choi2018stargan}, F-GAN \cite{palsson2018generative}, Age-cGAN \cite{antipov2017face}, IPCGANs \cite{wang2018face}, BeautyGAN \cite{li2018beautygan}, PairedCycleGAN \cite{chang2018pairedcyclegan}, G2-GAN \cite{song2018geometry}, GANimation \cite{pumarola2018ganimation}, GC-GAN \cite{qiao2018geometry}, ExprGAN \cite{ding2018exprgan}, DR-GAN \cite{tran2017disentangled}, CAPG-GAN \cite{hu2018pose}, UV-GAN \cite{deng2018uv}, CVAE-GAN \cite{bao2017cvae}, RenderGAN \cite{sixt2018rendergan}, DA-GAN \cite{zhao20183d, zhao2017dual}, TP-GAN \cite{huang2017beyond}, SimGAN \cite{shrivastava2017learning}, FF-GAN \cite{yin2017towards}, GP-GAN \cite{di2018gp}, and so on.

\subsubsection{Flow-based Generative Models}\label{sec:transmethod:generative:flow}

In addition to autoregressive models and VAEs, Flow-based generative models are likelihood-based generative methods as well, which were first described in NICE~\cite{dinh2014nice}. In order to model complex high-dimensional densities, they first map the data to a latent space where the distribution is easy to model. The mapping is performed through a non-linear deterministic transformation which can decomposed into a sequence of transformations and is invertible. Let $\textbf{x}$ be a high-dimensional random vector with unknown distribution, $\textbf{z}$ be the latent variable, the relationship between the $\textbf{x}$ and $\textbf{z}$ can be written as:

\begin{equation}
{\bf x}\stackrel{f_1}{\longleftrightarrow}h_1\stackrel{f_2}{\longleftrightarrow}h_2{\cdots}\stackrel{f_k}{\longleftrightarrow}{\bf z},
\end{equation}

where $\textbf{z}=f(\textbf{x})$ and $f=f_1{\circ}f_2{\circ}{\cdots}{\circ}f_k$. Such a sequence of invertible transformations is called a flow.

Flow-based generative models have not gained much attention so far. In fact, they have exact latent-variable inference and log-likelihood evaluation comparing to VAEs and GANs, and they are able to perform efficient inference and synthesis comparing to autoregressive models~\cite{kingma2018glow}. Kingma and Dhariwal proposed Glow~\cite{kingma2018glow}, a simple type of generative flow using an invertible $1{\times}1$ convolution. They demonstrated the model's ability in synthesizing high-resolution face images through a series of experiments for image generation, interpolation, and semantic manipulation. Although the results still had a gap with GANs, they showed significant improvement against previous flow-based generative models. Grover et al. introduced Flow-GAN~\cite{grover2018flow}, a generative adversarial network with a normalizing flow generator, with the purpose of bridging the gap between high-quality generated samples and ill-defined likelihood for GANs. It transformed the prior noise density into a model density through a sequence of invertible transformations, so the exact likelihoods could be tractably evaluated.

\subsubsection{Generative Models Comparison}\label{sec:transmethod:generative:comparison}

Each generative model has its pros and cons. Accurate likelihood evaluation and sampling are tractable in autoregressive models. They have given the best log likelihoods so far and have stable training process. However, they are less effective during sampling and the sequential generation process is slow. Variational autoencoders allow us to perform both learning and efficient inference in sophisticated probabilistic graphical models with approximate latent variables. Anyway, the likelihood is intractable to compute, and the variational lower bound to optimize for learning the model is not as exact as that in autoregressive models. Meanwhile, their generated samples tend to be blurry and of lower quality compared to GANs. GANs can generate sharp images, and there is no Markov chain or approx networks involved during sampling. However, they cannot provide explicit density. This makes it more challenging for quantitative evaluations~\cite{grover2018flow}, and also more difficult to optimize due to unstable training dynamics.

In order to improve the performance of generative models, many efforts have been made. For example, some works modified the architecture of these models to gain better characters, such as Gated PixelCNN \cite{van2016conditional}, CVAE \cite{sohn2015learning}, and DCGAN \cite{radford2015unsupervised}. Some works try to combine different models in one generative framework. An example of the combination of VAEs and PixelCNN is PixelVAE~\cite{gulrajani2016pixelvae}. It is a VAE model with an autoregressive decoder based on PixelCNN, whereas it has fewer autoregressive layers than PixelCNN and learns more compressed latent representations than standard VAE. Perarnau et al.~\cite{perarnau2016invertible} combined an encoder with a cGAN into IcGAN (Invertible cGAN), which enabled image generation with deterministic modification.~\cite{makhzani2015adversarial} and~\cite{larsen2015autoencoding} proposed similar ideas by combining VAE with GANs, and introduced AAE (adversarial autoencoder) and VAE/GAN, which could generate photorealistic images while keeping training stale. What's more, Zhang et al.~\cite{zhang2017age} designed CAAE (conditional adversarial autoencoder) for face age progression and regression, Zhou et al.~\cite{zhou2017photorealistic} presented CDAAE (conditional difference adversarial autoencoder) for facial expression synthesis. Bao et al.~\cite{bao2017cvae} proposed CVAE-GAN, a conditional variational generative adversarial network capable of generating images of fine-grained object categories, such as faces of a specific person.

\begin{figure}[htbp]
\centerline{\includegraphics[width=0.45\textwidth]{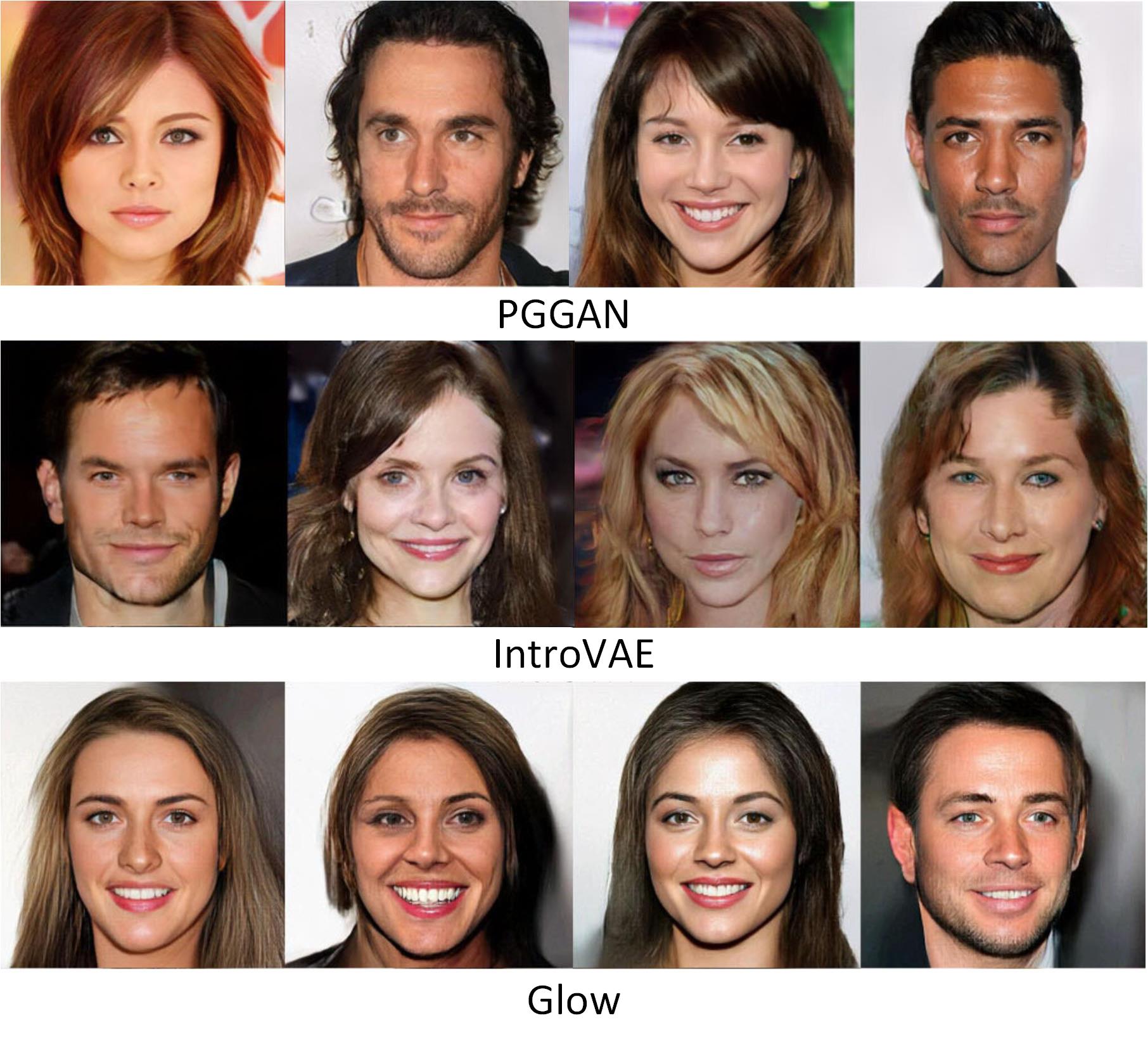}}
\caption{Samples from PGGAN \cite{karras2017progressive}, IntroVAE \cite{huang2018introvae}, and Glow \cite{kingma2018glow}}
\label{GenCom}
\end{figure}

Fig.~\ref{GenCom} presents some samples of high-resolution images generated by PGGAN~\cite{karras2017progressive}, IntroVAE~\cite{huang2018introvae}, and Glow~\cite{kingma2018glow}, which are the highest level representation of GANs, VAE, and Flow-based generative models at present from our view. Through a comparison of these images, it can be seen that GANs can synthesize more realistic images. The samples from PGGAN are more natural in terms of face shape, expression, hair, eyes, and lighting. However, they still have defects in some places, such as the asymmetry of color and shape at the areas of cloth and earrings. The generated faces by IntroVAE are not sufficiently "beautiful", which may be caused by the unnatural eyebrow, wrinkles, and others. Glow synthesizes images in a painting style, which is reflected clearly by the hairline and lighting.

\subsection{Augmented Reality}\label{sec:transmethod:ar}

\begin{figure}[htbp]
\centerline{\includegraphics[scale=0.25]{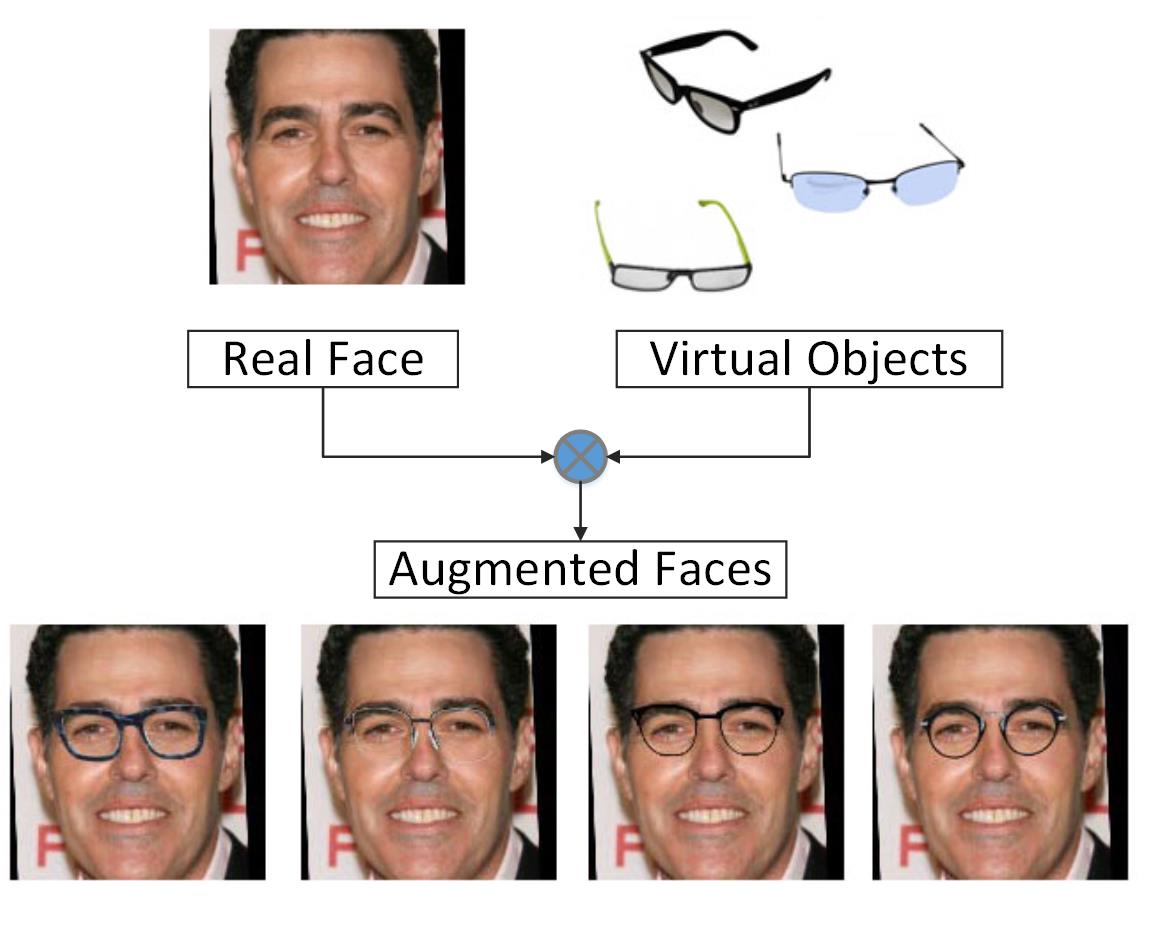}}
\caption{Applying AR for data augmentation. The real face and augmented faces are from LiteOn \cite{hsu2018investigating}.}
\label{ARFig}
\end{figure}

Augmented Reality (AR) is a technique which supplements the real word with virtual (computer-generated) objects that appear to coexist in the same space~\cite{van2010asurvey}. It allows a seamless fusion of virtual elements and real world, like fusing a real desk with a virtual lamp, or trying virtual glasses on a real face(Fig.~\ref{ARFig}). The application of AR technology can expand the scale of training data from the following aspects: supplementing the missing elements in the real scene, providing precise and detailed virtual elements, and increasing the data diversity.

In order to improve the performance of eyeglasses face recognition, Guo et al.~\cite{guo2018face} synthesized such images by reconstructing 3D face models and fitting 3D eyeglasses based on anchor points. The experiments on the real face dataset validated that their synthesized data had expected improvement in face recognition. LiteOn~\cite{hsu2018investigating} also generated augmented face images with eyeglasses to increase the accuracy of face recognition (see Fig. \ref{ARFig}). In fact, many AR techniques can be adopted for face data augmentation, such as the virtual mirror~\cite{kitanovski2011augmented} proposed for facial geometric alteration, the magic mirror~\cite{javornik2016revealing} used for makeup or accessories try-on, the Beauty e-Experts system~\cite{liu2014wow} designed for hairstyle and facial makeup recommendation and synthesis, the virtual glasses try-on system~\cite{azevedo2016augmented}, etc. More experiences of data augmentation by AR can be found in ~\cite{dosovitskiy2015flownet}, ~\cite{menze2015object}, and ~\cite{alhaija2017augmented}.

\subsection{Auto Augmentation}\label{sec:transmethod:auto}

Different augmentation strategies shoule be applied to different tasks. In order to automatically find the most appropriate scheme to enrich the training set and obtain an optimal result, some auto augmentation methods were proposed. For example,~\cite{wang2017effectiveness} applied an Augmentation Network (AugNet) to learn the best augmentation for improving a classifier, \cite{lemley2017smart} presented Smart Augmentation by creating one or multiple networks to learn the suitable augmentation for the given class of input data during the training of the target network. Different from the above works, \cite{cubuk2018autoaugment} created a search space for data augmentation policies, and used reinforcement learning as the search algorithm to find the best operations of data augmentation on the dataset of interest. During the training of the target network, it used a controller to predict augmentation decisions and evaluated them on a validation set in order to produce reward signals for the training of the controller. Remarkably, the current auto augmentation techniques are usually designed for simple augmentation operations, such as rotation, translation, and cropping.

\section{Evaluation Metrics}\label{sec:evaluation}

The two main methods of evaluation are the qualitative evaluation and quantitative evaluation. For qualitative evaluation, the authors directly present the readers or interviewers with generated images and let them judge the quality. The quantitative evaluation is usually based on some statistical methods. It provides quantifiable and explicit results. Most of the time, both qualitative and quantitative methods are employed together to provide adequate information for the evaluation, and suitable evaluation metrics should be applied since face data augmentation includes different transformation types with different purpose.

The frequently used metrics include the accuracy and error rate, distance measurement, Inception Score (IS)~\cite{salimans2016improved}, and Fr\'{e}chet Inception Distance (FID)~\cite{heusel2017gans}, which are introduced respectively as follows.

\emph{Accuracy and error rate} are the most commonly used measurements for classification and prediction, which are calculated on the numbers of positive and negative samples. Assume the numbers of positive samples and negative samples are $N_p$ and $N_n$, and the numbers of correct predictions for the positive and negative are $t_p$ and $t_n$ respectively. The accuracy rate is defined as $Accuracy Rate=((t_{p}+t_{n})/(N_{p}+N_{n}))$. The error rate is $Error Rate=1-Accuracy Rate$. However, the above equations only work for balanced data, which means the positive samples have equal number with the negative. Otherwise, it should adopt the balanced accuracy and error rate, which are defined as $Balanced Accuracy Rate=1/2(t_{p}/N_{p}+t_{n}/N_{n})$, and $Balabced Error Rate=1-Balanced Accuracy Rate$.

\emph{Distance measurement} can be used in a wide range of scenarios. For example, the L1 norm and L2 norm are usually adopted to calculate the color distance and spatial distance, the Peak Signal to Noise Ratio (PSNR) is used to measure pixel differences, and the distance of two probability distributions can be measured by the KL-divergence or Fr\'{e}chet distance. Usually, the distance metrics are used as the basis of other metrics.

\emph{Inception Score} measures the performance of a generative model from the quality of the generated images and their diversity. It is computed based on the relative entropy of two probability distributions which are relative to the output of a pretrained classification network. The IS is defined as $\rm{exp}$$(\mathbb{E}_{x{\thicksim}p_g}D_{KL}(p(y|x)||p(y)))$, where $x$ denotes the generated image, y is the class label output by the pretrained network, and $p_g$ represents the generated distribution. The principle of IS is that the images generated by a better generative model should have a conditional label distribution $p(y|x)$ with low entropy (which means the generated images are highly predictable) and a marginal distribution $p(y)=\int p(y|x=G(z)) dz$ with high entropy (which means the generated images are diverse).

\emph{Fr\'{e}chet Inception Distance} applies an inception network to extract image feature, and models the feature distributions of the generated and real images as two multivariate Gaussian distributions. The quality and diversity of the generated images are measured by calculating the Fr\'{e}chet distance of the two distributions, which can be expressed as $\rm{FID}$$(x,g)=||{\mu}_x-{\mu}_g||^2_2+\rm{Tr}({\sum}_x+{\sum}_g-2({\sum}_x{\sum}_g)^{\frac{1}{2}})$, where $({\mu}_x, {\sum}_x)$ and $({\mu}_g, {\sum}_g)$ are the mean and covariance of the two Gaussian distributions correlated with the real images and the generated images. In comparison with IS, FID is more robust to noise and more sensitive to intra-class mode collapse~\cite{huang2018introduction}.

Actually, it is difficult to give a totally fair comparison of different face augmentation methods with an uniform criteria, as they usually focus on diverse problems and are designed for different applications. For example, the image quality and diversity are the main concerns for image generation, so Inception Score and FID are the most widely adopted metrics. Whereas,
for conditional image generation, the generation direction or the generated images' domain is also important. Furthermore, if the work focuses on identity preservation, a face recognition or verification test is necessary. Usually, the importance of each component inside the framework is evaluated through ablation study. If the effect of a data augmentation method on a specific task is desired to be evaluated, a direct way is to compare the task execution result with and without the augmented data. One notable thing is that the task execution is based on specific algorithms and datasets, so it is impossible to evaluate over all possible algorithms and datasets.

\section{Challenges and Opportunities}\label{sec:challenges}

Face data augmentation is effective on enlarging the limited training dataset and improving the performance of face related detection and recognition tasks. Despite the promising performance achieved by massive existing works, there are still some issues to be tackled. In this section, we point out some challenges and interesting directions for the future research of face data augmentation.

\subsection{Identity Preserving}\label{sec:challenges:identity}

Although many facial attributes are transformed during the data augmentation procedure, the identity which is the most important class label for face recognition and verification, is usually expected to preserve in most cases. However, it still remains difficult in conditional face generation~\cite{shen2018facefeat}.

So far, the most commonly used approach in existing works is the introduction of an identity loss during network training. For example,~\cite{cao2018learning} added a perceptual identity loss to integrate domain knowledge of identity into the network, which is similar to~\cite{huang2017beyond, wang2018face, li2016deep, li2016convolutional, song2018geometry}. They employed the method proposed in~\cite{johnson2016perceptual} to calculate the perceptual loss, and extracted identity features through a face recognition module, e.g. VGG-Face~\cite{parkhi2015deep} or Light CNN~\cite{wu2015lightened}. In addition, the works~\cite{shen2018facefeat, bao2018towards, antipov2017face, wiles2018x2face} limited the identity feature distance (Manhattan distance or Euclidean distance) to decrease the identity loss in the process of face image generation.~\cite{deng2018uv} defined a centre loss for identity, based on the activations after the average pooling layer of ResNet.

Another way of identity preservation is to use cycle consistency loss to supervise the identity variation. For example, PairedCycleGAN~\cite{chang2018pairedcyclegan} contained an asymmetric makeup transfer framework for makeup applying and removal, where the output of the two asymmetric functions should get back the input. In \cite{palsson2018generative}, the cycle consistency loss for identity preserving is defined by comparing the original face with the aging-and-rejuvenating result. Some work applied both the perceptual loss and cycle consistency loss for facial identity preservation, such as~\cite{li2018beautygan}.

In addition to the methods mentioned above, some researchers adopted specific networks to supervise identity changes.~\cite{tran2017disentangled} modified the classical discriminator of GANs. In their DR-GAN for face pose synthesizing, the discriminator was trained to not only distinguish real and synthetic images, but also predict the identity and pose of a face. However, their multi-task discriminator classified all the synthesized faces as one class. In contrast, Shen et al.~\cite{shen2018faceid} proposed a three-player GAN, where the face classifier was treated as the third layer and competed with the generator. Moreover, their classifier differentiated the real and synthesized domains by assigning them different identity labels. The author claimed that previous methods cannot satisfy the requirement of identity preservation because they only tried to push real and synthesized domains close, but neglected how close they were.

Despite the efforts have been made, identity preservation is still a challenging problem which has not been completely solved. In fact, identity is presented by various facial attributes. Therefore, it is important to maintain the identity-relative features when changing other attributes, while this remains difficult for an end-to-end network.

\subsection{Disentangled Representation}\label{sec:challenges:disentangle}

Disentangled representation can improve the performance of neural networks in conditional generation by adjusting corresponding factors while keeping other attributes fixed. It enables a better control of the network output. For this purpose,~\cite{shu2017neural} introduced a special generative adversarial network by encoding the image formation and shading processes into network layers. In consequence, the network could infer a face-specific disentangled representation of intrinsic face properties (shape, albedo and lighting), and allowed for semantically relevant facial editing.~\cite{qiao2018geometry} used separate paths to learn the geometry expression feature and image identity feature in a disentangled manner.~\cite{bao2018towards} disentangled the identity and other attributes of faces by introducing an identity network and an attribute network to encode the identity and attributes into separate vectors before importing them to the generator.~\cite{shen2018facefeat} proposed a two-stage approach for face synthesis. It produced disentangled facial features from random noises using a series of feature generators in the first stage, and decoded these features into synthetic images through an image generator in the second stage.~\cite{chen2018texture} disentangled the texture and deformation of the input images by adopting the Intrinsic DAE (Deforming-Autoencoder) model~\cite{shu2018deforming}. It transferred the input image to three physical image signals (shading, albedo, and deformation), and combined the shading and albedo to generate texture image. Liu et al.~\cite{liu2018disentangling} introduced a composite 3D face shape model composed of mean face shape, identity-sensitive difference, and identity-irrelevant difference. They disentangled the identity and non-identity features in the 3D face shape, and represented them with separate latent variables.

The encoder-decoder architecture is widely used for face editing by mapping the input image into a latent representation and reconstructing a new face with desired attribute.~\cite{tran2017disentangled} disentangled the pose variation from identity representation by inputting a separate pose code to the decoder, and validating the generated pose with the discriminator.~\cite{chen2016infogan} learned disentangled and interpretable representations for images in an entirely unsupervised manner by adding new objective to maximize the mutual information between small subsets of the latent variables and the observation. \cite{zhou2017genegan} proposed GeneGAN, whose encoder decomposed the input image to background feature and object feature.~\cite{xiao2017dna} constructed a DNA-like latent representation, in which different pieces of encodings controlled different visual attributes. Similarly,~\cite{lample2017fader} tried to learn a disentangled latent space for explicit attribute control by applying adversarial training in latent space instead of the pixel space. However, as argued in~\cite{he2017attgan}, the attribute-independent constraint on the latent representation was excessive, as it may restrict the capacity of the latent representation and cause information loss. Instead, they defined constraint for the generated images through an attribute classification for accurate facial attribute editing.

\subsection{Unsupervised Data Augmentation}\label{sec:challenges:unsupervised}

Collecting large amounts of images with certain attributes is a difficult task, which limits the application of data augmentation based on supervised learning. Therefore, semi-supervised and unsupervised methods are proposed to reduce the data demand of face generation.~\cite{wiles2018x2face} produced generated faces with the pose and expression of the driving frame without requiring expression or pose label, or coupled training data either. In the training stage, it extracted source frame and driving frame from a same video, so the generated and driving frames would match.~\cite{kim2017learning} implemented cross-domain translation without any explicitly paired data. In~\cite{zhang2017age}, only multiple faces with different ages were used, and no paired samples for training or labeled faces for test were needed. \cite{song2018dual} developed a dual conditional GANs (Dual cGANs) for face aging and rejuvenation without the requirement of sequential training data.~\cite{shen2017learning} adopted dual learning, which could transform images inversely and learn from each other. \cite{moniz2018unsupervised} inferred the depth of facial keypoints without using any ground-truth of depth information. \cite{kim2017unsupervised} and \cite{pumarola2018ganimation} proposed unsupervised strategies for visual attributes transfer and expression transfer. \cite{zhu2017unpaired} presented cycleGAN, which could be used in image-to-image translation with no need for paired training data. Then \cite{song2018geometry} employed cycleGAN to simultaneously perform expression generation and removal. \cite{chang2018pairedcyclegan} applied cycleGAN for makeup applying and removal. Recently, \cite{lu2018attribute} proposed conditional CycleGAN for conditional face generation, which was also an unsupervised learning method.

Both the supervised and unsupervised learning methods have their respective pros and cons. On one hand, unsupervised learning methods make the preparing of training data much easier. It has boosted the development and application of face data augmentation. On the other hand, without the help of appropriatly classified and labeled data, the learning process becomes more difficult and unstable, and the learned model is less accurate. In~\cite{zhu2017unpaired}, a lingering gap between the results achievable with paired training data and unpaired data was observed, and it was believed that this gap would be hard or even impossible to close in some cases. In order to make up the defects in training data, extra information should be injected, such as prior knowledge or expert knowledge. Actually, image translation in unsupervised setting is a highly ill-posed problem, since there exist an infinite set of joint distributions that can achieve the two domains translation~\cite{liu2017unsupervised}. Therefore, more effort should be made to lower the training difficulty if less training data is desired.

\subsection{Improvement of GANs}\label{sec:challenges:improvement}

In recent years, GANs have become one of the most popular methods in data generation. It is easy to use, and has created a lot of impressive results. However, it still suffers from problems like instable training and mode collapse. Efforts to improve the effectiveness of GANs have never been stopped. Besides the works introduced in Sect. \ref{sec:transmethod:generative:gan}, many researchers modified the loss functions for higher quality of the generated results. For example, \cite{li2016convolutional} combined identity loss with attribute loss for attribute-driven and identity-preserving human face generation. \cite{li2018beautygan} applied four types of losses, including the adversarial loss, cycle consistency loss, perceptual loss and makeup constrain loss, to guarantee the quality of the makeup transfer. \cite{choi2018stargan} used adversarial loss, domain loss and reconstruction loss for the training of their StarGAN. As mentioned in Sect. \ref{sec:challenges:identity}, many works adopted identity loss to preserve the identity of the generated face.

\begin{figure}[htbp]
\centerline{\includegraphics[scale=0.25]{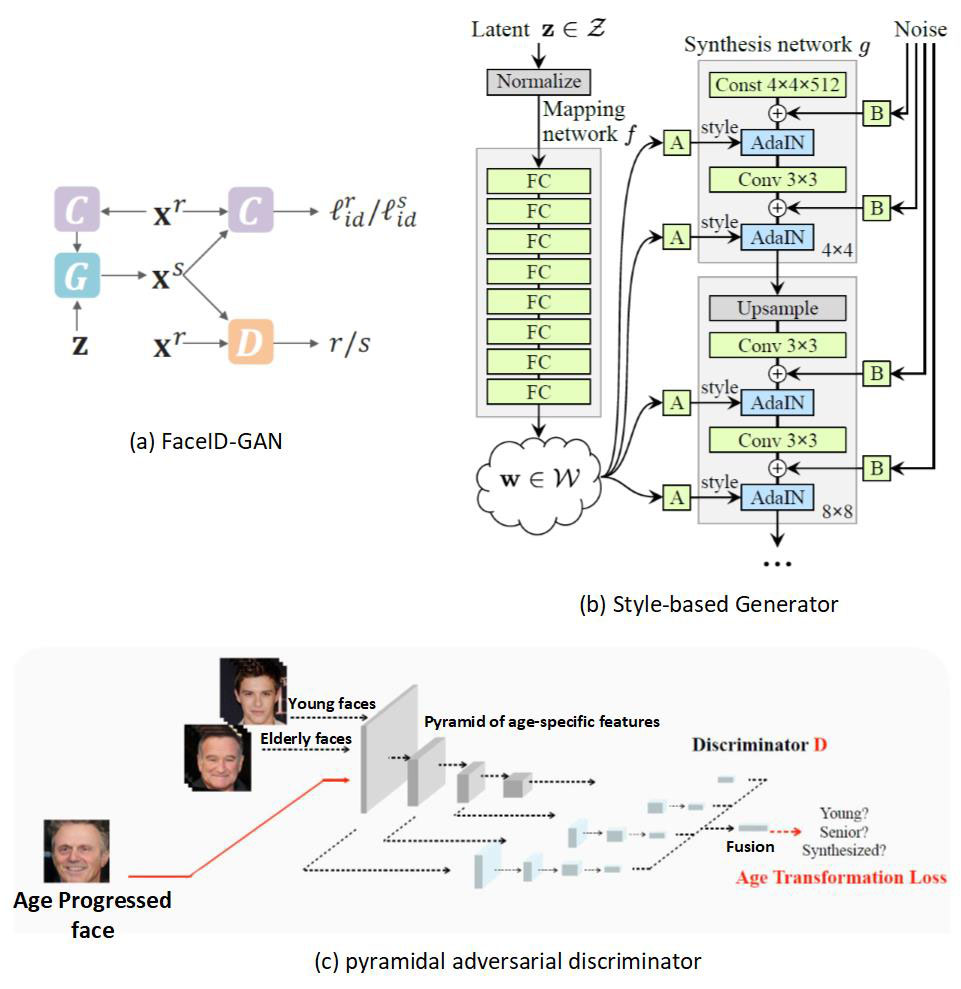}}
\caption{Improvement of GANs by (a)three-player GAN \cite{shen2018faceid}, (b)style-based generator \cite{karras2018style}, and (c)pyramidal adversarial discriminator \cite{yang2018learning}.}
\label{NewGan}
\end{figure}

Recently, some improvement for the architecture of the original GANs were presented.~\cite{shen2018faceid} proposed a symmetry three-player GAN -- FaceID-GAN. In contrast to the classical two-player game of most GANs, it applied a face identity classifier as the third player to distinguish the identities of the real and synthesized faces. The architecture of the FaceID-GAN is illustrated in Fig.~\ref{NewGan}-a, where $G$ represents the generator, $D$ is the discriminator, and $C$ is the classifier. The real image $x^{r}$ and the synthesized image $x^{s}$ are represented in the same feature space by using the same classifier $C$, in order to satisfy the principle of information symmetry and alleviate the training difficulty. The classifier $C$ is used to distinguish the identities of two domains, and collaborates with $D$ to compete generator $G$.~\cite{karras2018style} re-designed the generator architecture of GANs, and proposed a style-based generator. As shown in Fig.~\ref{NewGan}-b, they first map the input latent code $z$ to an intermediate latent space ${\mathcal W}$, which controls the generated styles through adaptive instance normalization (AdaIN) at each convolution layer. Another modification is the application of Gaussian noise input. It creates stochastic and localized variation to the generated images, leaving the high-level features such as identity intact. Inspired by the natural images that exhibit multi-scale characteristics along the hierarchical architecture,~\cite{yang2018learning} proposed a pyramid architecture for the discriminator of GANs (Fig. \ref{NewGan}-c). The pyramid faical feature representations are jointly estimated by $D$ at multiple scales, which handles face generation in a fine-gained way. Their evaluation result demonstrated that the pyramid structure advanced the generation of aged faces by making them more natural and possessing more face details.

Besides the above mentioned improvement,~\cite{banerjee2017srefi} proposed MAD-GAN (Multi-Agent Diverse GAN), which incorporated multiple generators and one discriminator. Gu et al.~\cite{gu2017differential} proposed a new differential discriminator and a network architecture, Differential Generative Adversarial Networks (D-GAN), in order to approximate the face manifold for non-linear facial variations with small amount of training data. Kossaifi et al.~\cite{kossaifi2018gagan} presented a method to incorporate geometry information into face image generation, and introduced the Geometry-Aware Generative Adversarial Networks (GAGAN). Juefei-Xu et al.~\cite{juefei2018rankgan} introduced a stage-wise learning paradigm for GANs that ranked multiple stages of generators by comparing the margin-based ranking loss of the generated samples. Tian et al.~\cite{tian2018cr} proposed a two-pathway (generation path + reconstruction path) framework, CR-GAN, to develop the widely used single-pathway (encoder-decoder-discriminator) network. The two pathways combined with self-supervised learning can learn complete representations in the embedding space, and produce high-quality image generations from unseen data in wild conditions.

\section{Discussion}\label{sec:discussion}

The lack of labeled samples has been a common problem for researchers and engineers working with deep learning. Undoubtedly, Data Augmentation is an effective tool for solving this problem and has been widely used in various tasks. Among these tasks, face data augmentation is more complicated and challenging than others. Various methods have been proposed to transform a real face image to a new type, such as pose transfer, hairstyle transfer, expression transfer, makeup transfer, and age transfer. Meanwhile, the simulated virtual faces can also be enhanced to be as realistic as the real ones. All these augmentation methods can be used to increase the variation of the training data and improve the robustness of the learned model.

Image processing is a traditional but powerful method for image transformation, which has been widely used in geometric and photometric transformations of generic data. It is more suitable for transforming the entire image in a uniform manner, rather than changing the facial attributes that need to transform a specific part or property of faces. Model-based method is intuitively suitable for virtual face generation. With the reconstructed face model, it is easy to modify the shape, texture, illumination and expression. However, it remains difficult to reconstruct a complete and precise face model from a single 2D image, and synthesizing a virtual face to be really realistic is still computationally expensive. Therefore, several realism enhancement algorithms were proposed. With the rapid development of deep learning, learning-based method has become more and more popular. Although challenges such as identity preservation still exist, many remarkable achievements have been made.

So far, deep neural networks have been able to generate very photorealistic images, which are even difficult to distinguish by human. However, some limitations exist in the controllability of data generation, and the diversity of generated data. One widely recognized disadvantage of neural networks is their "black box" nature, which means we don't know how to operate the intermediate output to modify some specific facial features. There have been some works try to infer the meanings of the latent vectors through the generated images~\cite{guan2018tlgan}. But this additional operation has an upper limit of capability, which cannot disentangle facial features in the latent space completely. About image diversity, the facial appearance variation in the real world is unmeasurable, while the variability is limited with regard to synthetic data. If the face changes too much, it will be more difficult to preserve the identity.

\section{Conclusion}\label{sec:conclusion}

In this paper, we give a systematic review of face data augmentation and show the wide application and huge potential for various face detection and recognition tasks. We start with an introduction about the background and the concept of face data augmentation, which are followed by a brief review of related work. Then, a detailed description of different transformations is presented to show the types supported for face data augmentation. Next, we present the commonly used methods of face data augmentation by introducing their principles and giving comparisons. The evaluation metrics for these methods are introduced subsequently. Finally, we discuss the challenges and opportunities. We hope this survey can give the beginners a general understanding of this filed, and give the related researchers the insight on future studies.





\bibliography{ref}{}
\bibliographystyle{IEEEtran}

\onecolumn
\appendix
A summery of recent works on face data augmentation is illustrated in the following table. It summarizes these works from the transformation type, method, and the evaluations they performed to test the capability of their algorithms for data augmentation. There is one notable thing that we only label the transformation types explicitly mentioned in original papers. Maybe the methods could be used to do other transformations but the authors did not mention in their original paper.
\\

\begin{center}
\scriptsize
\tablehead{
\hline
\vspace{0.45cm}\centering
\multirow{2}*{Work}&
\vspace{0.45cm}\centering
\multirow{2}*{Alias}&
\multicolumn{7}{c|}{Supported Transformation Types}&
\vspace{0.45cm}\centering
\multirow{2}*{Method}&
\vspace{0.4cm}
\multirow{2}*{
\begin{minipage}[c]{0.07\textwidth}
\centering Evaluated \newline Task
\end{minipage}} \\
\cline{3-9}
& &HairStyle&Makeup&Accessory&Pose&Expression&Age&Others& & \\
[0.3cm] \hline
}
\tabletail{
\hline
}
\topcaption{Summary of recent works}
\begin{supertabular*}{0.95\textwidth}{@{\extracolsep{\fill}}|m{1.3cm}<{\centering}|m{1.7cm}<{\centering}|ccccccm{1.2cm}<{\centering}|m{1.7cm}<{\centering}|m{1.7cm}<{\centering}|}
\vspace{0.1cm}\centering
Zhu et al.\cite{zhu2018emotion}& -- & & & & & $\surd$ & & & GANs-based & Emotion Classification \\
\hline
\vspace{0.1cm}\centering
Feng et al.\cite{feng2017dynamic}& -- & & & & $\surd$ & & & & 2D Model-based & -- \\
\hline
\vspace{0.1cm}\centering
Zhu et al.\cite{zhu2016face}& -- & & & & $\surd$ & & & & 3D Model-based & Face Alignment \\
\hline
\vspace{0.1cm}\centering
Masi et al.\cite{masi2016pose}& -- & & & & $\surd$ & & & & 3D Model-based & Face Recognition \\
\hline
\vspace{0.1cm}\centering
Hassner et al.\cite{hassner2015effective}& -- & & & & $\surd$ & & & & 3D Model-based & -- \\
\hline
\vspace{0.1cm}\centering
Bao et al.\cite{bao2018towards}& -- & & & & $\surd$ & $\surd$ & & Illumination & GANs-based & -- \\
\hline
\vspace{0.1cm}\centering
Kim et al.\cite{kim2017learning}& DiscoGAN & $\surd$ & & $\surd$ & $\surd$ & & & Gender & GANs-based & -- \\
\hline
\vspace{0.1cm}\centering
Choi et al.\cite{choi2018stargan}& StarGAN & $\surd$ & & & & $\surd$ & $\surd$ & Gender Skin & GANs-based & -- \\
\hline
\vspace{0.1cm}\centering
Isola et al.\cite{isola2017image}& pix2pix & & & & & & & Background & GANs-based & -- \\
\hline
\vspace{0.1cm}\centering
Liu et al.\cite{liu2017unsupervised}& -- & $\surd$ & & $\surd$ & & $\surd$ & & Goatee & Generative-based & -- \\
\hline
\vspace{0.1cm}\centering
Liu et al.\cite{liu2016coupled}& CoGAN & $\surd$ & & $\surd$ & & $\surd$ & & & GANs-based & -- \\
\hline
\vspace{0.1cm}\centering
Palsson et al.\cite{palsson2018generative}& Group-GAN FA-GAN F-GAN & & & & & & $\surd$ & & GANs-based & -- \\
\hline
\vspace{0.1cm}\centering
Zhang et al.\cite{zhang2017age}& CAAE & & & & & & $\surd$ & & Generative-based & -- \\
\hline
\vspace{0.1cm}\centering
Antipov et al.\cite{antipov2017face}& Age-cGAN & & & & & & $\surd$ & & GANs-based & -- \\
\hline
\vspace{0.1cm}\centering
Shen et al.\cite{shen2017learning}& -- & & & $\surd$ & & $\surd$ & $\surd$ & Beard Gender & GANs-based & -- \\
\hline
\vspace{0.1cm}\centering
Kim et al.\cite{kim2017unsupervised}& -- & $\surd$ & & & $\surd$ & $\surd$ & & & GANs-based & -- \\
\hline
\vspace{0.1cm}\centering
Masi et al.\cite{masi2016we}& -- & & & & $\surd$ & $\surd$ & & Shape & 3D Model-based & Face Recognition \\
\hline
\vspace{0.1cm}\centering
Lv et al.\cite{lv2017data}& -- & $\surd$ & & $\surd$ & $\surd$ & & & Illumination \newline Landmark-Perturbation & Image-Process \newline 3D Model-based & Face Recognition \\
\hline
\vspace{0.1cm}\centering
Xie et al.\cite{xie2018facial}& -- & & & & & $\surd$ & & & Image-Process & --\\
\hline
\vspace{0.1cm}\centering
Thies et al.\cite{thies2015real}& -- & & & & & $\surd$ & & & 3D Model-based & -- \\
\hline
\vspace{0.1cm}\centering
Guo et al.\cite{guo2017facenet3d}& -- & & & & $\surd$ & $\surd$ & & & 3D Model-based & -- \\
\hline
\vspace{0.1cm}\centering
Kim et al.\cite{kim2017deep}& -- & & & & $\surd$ & $\surd$ & & Occlusion & 3D Model-based & 3D Face Recognition \\
\hline
\vspace{0.1cm}\centering
Zhou et al.\cite{zhou2017photorealistic}& CDAAE & & & & & $\surd$ & & & Generative-based & -- \\
\hline
\vspace{0.1cm}\centering
Yeh et al.\cite{yeh2016semantic}& FVAE & & & & & $\surd$ & & & Generative-based & -- \\
\hline
\vspace{0.1cm}\centering
Li et al.\cite{li2016deep}& DIAT & & & $\surd$ & & $\surd$ & $\surd$ & Gender & GANs-based & -- \\
\hline
\vspace{0.1cm}\centering
Oord et al.\cite{van2016conditional}& -- & & & & $\surd$ & $\surd$ & & Illumination & Generative-based & -- \\
\hline
\vspace{0.1cm}\centering
Zhang et al.\cite{zhang2018joint}& -- & & & & $\surd$ & $\surd$ & & & GANs-based & Expression Recognition \\
\hline
\vspace{0.1cm}\centering
He et al.\cite{he2017attgan}& AttGAN & $\surd$ & & $\surd$ & & $\surd$ & $\surd$ & Beard Eyebrows Gender Skin & GANs-based & -- \\
\hline
\vspace{0.1cm}\centering
Pumarola et al.\cite{pumarola2018ganimation}& GANimation & & & & & $\surd$ & & & GANs-based & -- \\
\hline
\vspace{0.1cm}\centering
Song et al.\cite{song2018geometry}& G2-GAN & & & & & $\surd$ & & & GANs-based & -- \\
\hline
\vspace{0.1cm}\centering
Qiao et al.\cite{qiao2018geometry}& GC-GAN & & & & & $\surd$ & & & GANs-based & -- \\
\hline
\vspace{0.1cm}\centering
Ding et al.\cite{ding2018exprgan}& ExprGAN & & & & & $\surd$ & & & GANs-based & Expression Classification\\
\hline
\vspace{0.1cm}\centering
Guo et al.\cite{guo2009digital}& -- & & $\surd$ & & & & & & Image-Process& -- \\
\hline
\vspace{0.1cm}\centering
Oo et al.\cite{oo2016digital}& -- & & $\surd$ & & & & & & Image-Process & -- \\
\hline
\vspace{0.1cm}\centering
Lee et al.\cite{lee2016anew}& -- & & $\surd$ & & & & & & Image-Process & -- \\
\hline
\vspace{0.1cm}\centering
Li et al.\cite{li2018beautygan}& BeautyGAN & & $\surd$ & & & & & & GANs-based & -- \\
\hline
\vspace{0.1cm}\centering
Chang et al.\cite{chang2018pairedcyclegan}& -- & & $\surd$ & & & & & & GANs-based & -- \\
\hline
\vspace{0.1cm}\centering
Huang et al.\cite{huang2017beyond}& TP-GAN & & & & $\surd$ & & & & GANs-based & -- \\
\hline
\vspace{0.1cm}\centering
Yin et al.\cite{yin2017towards}& FF-GAN & & & & $\surd$ & & & & GANs-based & -- \\
\hline
\vspace{0.1cm}\centering
Tran et al.\cite{tran2017disentangled}& DR-GAN & & & & $\surd$ & & & & GANs-based & -- \\
\hline
\vspace{0.1cm}\centering
Wiles et al.\cite{wiles2018x2face}& X2Face & & & & $\surd$ & $\surd$ & & & Generative-based & -- \\
\hline
\vspace{0.1cm}\centering
Crispell et al.\cite{crispell2017dataset}& -- & & & & $\surd$ & & & Illumination & 3D Model-based & Face Recognition \\
\hline
\vspace{0.1cm}\centering
Kulkarni et al.\cite{kulkarni2015deep}& DC-IGN & & & & $\surd$ & & & Illumination & VAEs-based & -- \\
\hline
\vspace{0.1cm}\centering
Zhao et al.\cite{zhao20183d, zhao2017dual}& DA-GAN & & & & $\surd$ & & & & 3D Model \& GANs & Face Recognition \\
\hline
\vspace{0.1cm}\centering
Kemelmacher et al.\cite{kemelmacher2016transfiguring}& -- & $\surd$ & & & & & $\surd$ & Beard & Image-Process & -- \\
\hline
\vspace{0.1cm}\centering
Chen et al.\cite{chen2016infogan}& InfoGAN & $\surd$ & & $\surd$ & $\surd$ & $\surd$ & & Illumination Shape & GANs-based & -- \\
\hline
\vspace{0.1cm}\centering
Wang et al.\cite{wang2018face}& IPCGANs & & & & & & $\surd$ & & GANs-based & Face Recognition \\
\hline
\vspace{0.1cm}\centering
Hu et al.\cite{hu2018pose}& CAPG-GAN & & & & $\surd$ & & & & GANs-based & -- \\
\hline
\vspace{0.1cm}\centering
Deng et al.\cite{deng2018uv}& UV-GAN & & & & $\surd$ & & & & 3D Model \& GANs & Face Recognition \\
\hline
\vspace{0.1cm}\centering
Bao et al.\cite{bao2017cvae}& CVAE-GAN & & & & $\surd$ & $\surd$ & & & Generative-based & Face Recognition \\
\hline
\vspace{0.1cm}\centering
Gecer et al.\cite{gecer2018semi}& -- & & & & $\surd$ & $\surd$ & & Illumination & 3D Model \& GANs & Face Recognition \\
\hline
\vspace{0.1cm}\centering
Pandey et al.\cite{pandey2017variational}& CMMA & & & $\surd$ & & & & Beard Shape & Generative-based & -- \\
\hline
\vspace{0.1cm}\centering
Yan et al.\cite{yan2016attribute2image}& disCVAE & $\surd$ & & $\surd$ & & $\surd$ & $\surd$ & Gender & Generative-based & -- \\
\hline
\vspace{0.1cm}\centering
Kingma et al.\cite{kingma2018glow}& Glow & $\surd$ & & & & $\surd$ & $\surd$ & Skin Gender & Generative-based & -- \\
\hline
\vspace{0.1cm}\centering
Huang et al.\cite{huang2018introvae}& IntroVAE & & & & $\surd$ & & & Gender & Generative-based & -- \\
\hline
\vspace{0.1cm}\centering
Zhao et al.\cite{zhao2018look}& FSN & & & & & & $\surd$ & & GANs-based & -- \\
\hline
\vspace{0.1cm}\centering
Zhu et al.\cite{zhu2018facial}& CMAAE-OR & & & & & & $\surd$ & & GANs-based & -- \\
\hline
\vspace{0.1cm}\centering
Song et al.\cite{song2018dual}& Dual cGANs & & & & & & $\surd$ & & GANs-based & -- \\
\hline
\vspace{0.1cm}\centering
Gu et al.\cite{gu2017differential}& D-GAN & & & & $\surd$ & $\surd$ & & Illumination & GANs-based & Expression Classification \\
\hline
\vspace{0.1cm}\centering
Kossaifi et al.\cite{kossaifi2018gagan}& GAGAN & & & & $\surd$ & $\surd$ & & & GANs-based & -- \\
\hline
\vspace{0.1cm}\centering
Cao et al.\cite{cao2018load}& LB-GAN & & & & $\surd$ & & & & GANs-based & -- \\
\hline
\vspace{0.1cm}\centering
Guo et al.\cite{guo2018face}& -- & & & $\surd$ & & & & & Augmented Reality & Face Recognition \\
\hline
\vspace{0.1cm}\centering
Wu et al.\cite{wu2018reenactgan}& ReenactGAN & & & & & $\surd$ & & & GANs-based & -- \\
\hline
\vspace{0.1cm}\centering
Liu et al.\cite{liu2018attribute}& -- & & & & & & $\surd$ & & GANs-based & -- \\
\hline
\vspace{0.1cm}\centering
Pham et al.\cite{pham2018generative}& -- & & & & & $\surd$ & & & GANs-based & -- \\
\hline
\vspace{0.1cm}\centering
Sanchez et al.\cite{sanchez2018triple}& GANnotation & & & & $\surd$ & $\surd$ & & & GANs-based & -- \\
\hline
\vspace{0.1cm}\centering
Li et al.\cite{li2018global}& WaveletGLCA-GAN & & & & & & $\surd$ & & GANs-based & -- \\
\hline
\vspace{0.1cm}\centering
Tian et al.\cite{tian2018cr}& CR-GAN & & & & $\surd$ & & & & GANs-based & -- \\
\hline
\vspace{0.1cm}\centering
Chen et al.\cite{chen2018texture}& TDB-GAN & & & $\surd$ & & $\surd$ & $\surd$ & Skin Gender & GANs-based & -- \\
\hline
\vspace{0.1cm}\centering
Lample et al.\cite{lample2017fader}& Fader Networks & & & $\surd$ & & $\surd$ & $\surd$ & Gender & Generative-based & -- \\
\hline
\vspace{0.1cm}\centering
Xiao et al.\cite{xiao2017dna}& DNA-GAN & $\surd$ & & $\surd$ & & $\surd$ & & Illumination \newline Gender \newline Hat \newline Mustache & GANs-based & -- \\
\hline
\vspace{0.1cm}\centering
Zhou et al.\cite{zhou2017genegan}& GeneGAN & $\surd$ & & $\surd$ & & $\surd$ & & Illumination & GANs-based & -- \\
\hline
\end{supertabular*}
\label{tab3}
\end{center}
\twocolumn

\end{document}